\documentclass[10pt,nologo]{lumia}

\usepackage[round,authoryear]{natbib}
\bibpunct{(}{)}{;}{a}{,}{,}
\let\cite\citep
\usepackage{xspace}

\usepackage{algorithm}
\usepackage{algorithmic}

\usepackage{multirow}
\usepackage{adjustbox}
\usepackage{wrapfig}

\usepackage[nameinlink,capitalise]{cleveref}
\usepackage{bookmark}

\newcommand{\ourmethod}{{\fontfamily{lmtt}\selectfont \textbf{StreamFlow}}\xspace}
\definecolor{tblours}{gray}{0.92}
\definecolor{questionpurple}{HTML}{F2ECFA}
\definecolor{questionpurpleborder}{HTML}{C9B4E8}
\definecolor{citationblue}{HTML}{00AEEF}
\definecolor{refpurple}{HTML}{8E6AD8}
\hypersetup{linkcolor=refpurple,citecolor=citationblue}
\newcommand{\obsbox}[1]{%
    \begin{tcolorbox}[
        colback=questionpurple,
        colframe=questionpurpleborder,
        boxrule=0.6pt,
        arc=3pt,
        boxsep=0pt,
        left=7pt,right=7pt,top=4pt,bottom=4pt
    ]
        \centering #1
    \end{tcolorbox}%
}

\newcommand{\rightfigureblock}[4]{%
    \par\Needspace{12\baselineskip}%
    \noindent
    \begin{minipage}[t]{0.475\textwidth}
        \vspace{0pt}#1
    \end{minipage}\hfill
    \begin{minipage}[t]{0.495\textwidth}
        \vspace{0pt}\centering
        #2
        \captionsetup{hypcap=false}%
        \captionsetup[figure]{skip=5pt}%
        \captionof{figure}{#3}
        \label{#4}
    \end{minipage}%
    \par\medskip
}

\newcommand{\rightwrapfigure}[4][r]{%
    \par
    \begin{wrapfigure}{#1}{0.495\textwidth}
        \vspace{-0.8\baselineskip}
        \centering
        #2
        \captionsetup{skip=5pt}%
        \caption{#3}
        \label{#4}
        \vspace{0.25\baselineskip}
    \end{wrapfigure}%
}

\newcommand{\rightwrapfiguretight}[4][r]{%
    \par
    \begin{wrapfigure}{#1}{0.495\textwidth}
        \vspace{-0.8\baselineskip}
        \centering
        #2
        \captionsetup{skip=0pt}%
        \caption{#3}
        \label{#4}
        \vspace{0.25\baselineskip}
    \end{wrapfigure}%
}

\newcommand{\inlinewrapfigure}[4][r]{%
    \begin{wrapfigure}{#1}{0.495\textwidth}
        \vspace{-0.8\baselineskip}
        \centering
        #2
        \captionsetup{skip=5pt}%
        \caption{#3}
        \label{#4}
        \vspace{0.25\baselineskip}
    \end{wrapfigure}%
}

\newcommand{\rightwraptable}[4][r]{%
    \begin{wraptable}{#1}{0.495\textwidth}
        \vspace{-0.8\baselineskip}
        \centering
        \caption{#3}
        \label{#4}
        #2
        \vspace{0.25\baselineskip}
    \end{wraptable}%
}

\hypersetup{
    pdftitle={StreamFlow: Dynamic Memory Flows for Streaming Video Understanding},
    pdfauthor={Muxin Fu, Yifan Zhang, Wentao Zhang, Fangming Guo, Qian Chen, Guibin Zhang, Shuicheng Yan, and Bo An}
}

\title{StreamFlow: Dynamic Memory Flows for Streaming Video Understanding}
\setheadertitle{StreamFlow: Dynamic Memory Flows for Streaming Video Understanding}
\makeatletter
\renewcommand{\@author}{%
  {\Authfont
    Muxin Fu\textsuperscript{1,2,*}, Yifan Zhang\textsuperscript{3,*},
    Wentao Zhang\textsuperscript{2,*}, Fangming Guo\textsuperscript{1},
    Qian Chen\textsuperscript{1,4}\\
    Guibin Zhang\textsuperscript{5}, Shuicheng Yan\textsuperscript{5,$\dagger$},
    Bo An\textsuperscript{2,$\dagger$}\\[0.5em]
  }%
  {\Affilfont
    \textsuperscript{1}Tongji University,
    \textsuperscript{2}Nanyang Technological University,
    \textsuperscript{3}University of Michigan\\
    \textsuperscript{4}The Hong Kong University of Science and Technology,
    \textsuperscript{5}National University of Singapore\\
    \textsuperscript{*}Equal contribution,\quad
    \textsuperscript{$\dagger$}Corresponding authors
  }%
}
\makeatother
\date{}

\begin{document}

\begin{abstract}
Streaming video understanding requires multimodal large language models (MLLMs) to preserve relevant evidence from continuously evolving streams under strict causality and bounded memory. Yet existing paradigms remain limited: model-based methods require intrusive backbone updates, while memory-based methods expend substantial visual-encoding computation on temporally redundant content and rely on rigid access to visual history. To address these limitations, we introduce \ourmethod, an efficient visual memory framework that enables dynamic, on-demand access to historical visual information. \ourmethod combines a lightweight, dynamics-aware mid-term memory that filters temporal redundancy before visual encoding with a latent long-term memory that consolidates historical video content into visual latents accessible to subsequent reasoning. During generation, an attention-guided retrieval mechanism injects relevant visual latents when the model's reliance on visual evidence weakens. \ourmethod achieves state-of-the-art streaming video understanding performance, reaching $67.73\%$ overall accuracy on StreamingBench, while also delivering strong performance on offline long-video benchmarks. Relative to the vanilla setting, it improves the visual attention score (VAS) by $59.1\%$ while reducing end-to-end latency and peak memory by $50.4\%$ and $21.1\%$, respectively, enabling more visually grounded and efficient reasoning.
\end{abstract}
\maketitle

\section{Introduction}

\rightfigureblock{%
Recent years have witnessed remarkable advances in the capabilities of multimodal large language models (MLLMs) across a broad range of video understanding tasks~\citep{tang2025video,bai2025qwen3,team2026qwen3}.
However, emerging scenarios such as autonomous driving~\citep{fu2025orion}, embodied robotics~\citep{black2024pi_0} and computer-using agents~\citep{tan2024cradle} increasingly demand real-time understanding of continuously evolving visual streams.
This growing demand has shifted research attention from conventional offline video understanding toward the more challenging paradigm of streaming video understanding~\citep{zeng2026streamforest,qian2025dispider,huang2025online}.
}{%
\vspace{6.5pt}%
\includegraphics[width=\linewidth,trim=23bp 8bp 21bp 11bp,clip]{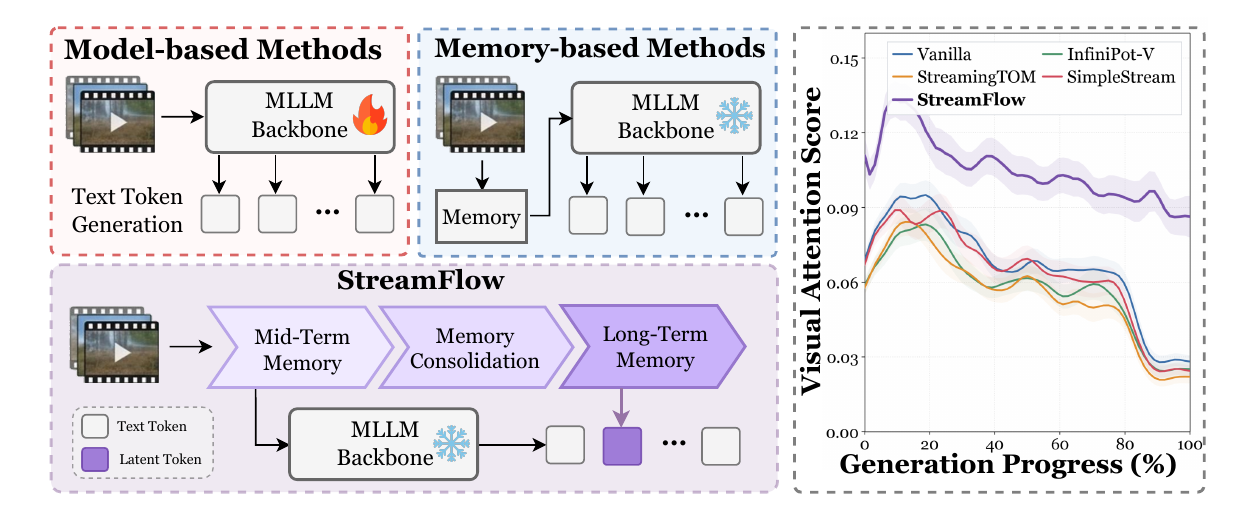}
}{Overview and motivation of \ourmethod. Left: comparison of model-based and memory-based methods with \ourmethod. Right: visual attention dynamics during autoregressive generation.}{fig:introduction}

Despite its importance, streaming video understanding remains challenging, requiring MLLMs to retain relevant evidence from unbounded streams under strict causality and resource constraints~\citep{lin2026streamingbench,qian2024streaming}.
Existing approaches can be broadly categorized into two paradigms, as illustrated in \Cref{fig:introduction} (left).
\textbf{(I) Model-based methods} fine-tune the MLLM backbone on streaming videos to endow the model with online perception~\citep{huang2025online,guan2026video,liu2026thinking}.
While this approach can yield substantial performance gains, adapting the full backbone incurs prohibitive training costs and risks catastrophic forgetting~\citep{zhou2024empirical,xie2025adadare}.
\textbf{(II) Memory-based methods}, by contrast, couple a frozen MLLM with an external memory that incrementally stores historical visual information~\citep{di2025streaming,ning2025livevlm,kim2026infinipot,chen2026streamingtom}.
This modular paradigm has gained growing attention, as it avoids full-backbone training and integrates seamlessly with pretrained MLLMs.

However, despite their efficiency, existing memory-based approaches still provide limited fine-grained management over visual information in two key aspects:
\textbf{(I) Redundant visual encoding:} existing approaches densely encode incoming frames before memory construction, spending substantial computation on static or predictable regions~\citep{tang2026onevision,shen2026simple,yao2025timechat}. \textbf{(II) Rigid memory access:} historical evidence is typically confined to a fixed visual prefix, incurring persistent context overhead and causing visual attention to decline during generation, as shown in \Cref{fig:introduction} (right), thereby increasing hallucination risk~\citep{xu2026more,huang2026persistent,yu2026vismem}.
\obsbox{\textit{How can we design a streaming memory that decides what visual information to encode and when to reactivate historical evidence during generation?}}

To address these limitations, we propose \ourmethod, a novel memory framework that efficiently enables dynamic, on-demand access to historical visual information. Specifically, \ourmethod comprises two complementary memory modules and an attention-guided injection mechanism: \textbf{(I) Dynamics-aware mid-term memory} identifies temporal changes directly from raw pixel differences and selectively encodes dynamic visual content, reducing redundant encoding while providing the MLLM with direct access to recent spatiotemporal information. \textbf{(II) Latent long-term memory} stores and consolidates earlier visual content as latent representations within a fixed capacity, keeping memory usage bounded as the video stream grows while making long-term visual information available for flexible access during generation. \textbf{(III) Attention-guided memory injection} monitors the MLLM's attention to visual information and injects visual latents retrieved from the long-term memory when needed. These components together form a continuous memory flow in which visual information is selectively encoded, consolidated over time, and dynamically reactivated when grounding weakens during generation.

Extensive evaluations across diverse video understanding benchmarks consistently demonstrate the effectiveness of \ourmethod in both streaming and offline settings. \textbf{(I) Strong Performance.} On StreamingBench, \ourmethod surpasses the previous best result by $4.63\%$. Its benefits extend to long-video understanding, outperforming leading streaming methods by $2.24\%$ and $8.22\%$ on MLVU and VideoMME, respectively. \textbf{(II) Improved Visual Grounding.} Mechanistic analysis shows that attention-guided injection improves the mean visual attention score (VAS) by $59.1\%$ relative to the vanilla setting, indicating that \ourmethod restores attention to historical visual evidence as generation proceeds. \textbf{(III) Greater Efficiency.} Relative to the vanilla setting, \ourmethod further reduces end-to-end latency and peak memory by $50.4\%$ and $21.1\%$, respectively.

In summary, our main contributions are as follows:

\begin{itemize}[leftmargin=1.6em]
    \item \textbf{Complementary Visual Memory.}
    We introduce mid-term and long-term memories that selectively encode evolving content and consolidate historical evidence within a fixed capacity.

    \item \textbf{Attention-Guided Injection.}
    We develop a visual attention score-guided mechanism that dynamically injects relevant visual latent memory during generation, mitigating visual attention drift.

    \item \textbf{Extensive Evaluation.}
    \ourmethod establishes state-of-the-art performance on streaming benchmarks while delivering strong offline performance, improved visual grounding, and greater inference efficiency.
\end{itemize}

\section{Related Work}

\paragraph{Streaming Video Understanding.}
Although video MLLMs have demonstrated strong performance in offline settings, streaming video understanding requires models to causally process continuously arriving visual inputs under strict latency and memory constraints~\citep{tang2025video}.
A major line of work equips MLLMs with streaming capabilities through training objectives designed for streaming and online instruction tuning~\citep{chen2024videollm,guan2026video,huang2025online,liu2026thinking},
often together with specialized designs for perception, computation, memory, and reasoning~\citep{he2024ma,qian2025dispider,song2024moviechat,wu2024videollm,zeng2026streamforest}.
However, in many such systems, the modules introduced for streaming and the MLLM backbone are jointly optimized to align their representations.
While effective, such joint optimization introduces substantial training costs and may compromise some of the general capabilities acquired during pretraining.
This motivates lightweight approaches that endow frozen MLLMs with streaming capabilities without modifying their pretrained parameters.

\paragraph{Streaming Video Memory.}
Memory mechanisms summarize unbounded video streams into compact representations, enabling MLLMs to process streaming visual inputs~\citep{meng2026watch,zou2024seconds}.
Existing methods can be characterized by their memory construction and utilization~\citep{shan2025cognitive,zhang2025memory}.
For memory construction, prior work compresses visual tokens or video KV caches under bounded memory budgets, often organizing historical information through recurrent, hierarchical, or multi-stage memory structures~\citep{chen2026scaling,chen2026streamingtom,kim2026infinipot,wang2025videollamb,xie2026fluxmem}.
For memory utilization, stored information is commonly retrieved from external caches at query time~\citep{di2025streaming,ning2025livevlm,chen2026streamingtom} or incorporated into the model through predefined access mechanisms~\citep{chen2026streamkv,liang2026oasis,zhang2026hermes}.
Representative approaches include ReKV~\citep{di2025streaming}, which stores historical video KV caches in an external memory, and TimeChat-Online~\citep{yao2025timechat}, which applies differential temporal token dropping to progressively condense streaming visual tokens. Despite these advances, many existing systems identify temporal redundancy only after visual encoding or expose historical memory through fixed or predetermined access patterns. In contrast, \ourmethod performs temporal filtering directly on raw frames before visual encoding and dynamically injects relevant long-term memory based on the model's reliance on visual information during generation.

\begin{figure*}[t]
    \centering
    \includegraphics[width=\linewidth,trim=0 10bp 0 0,clip]{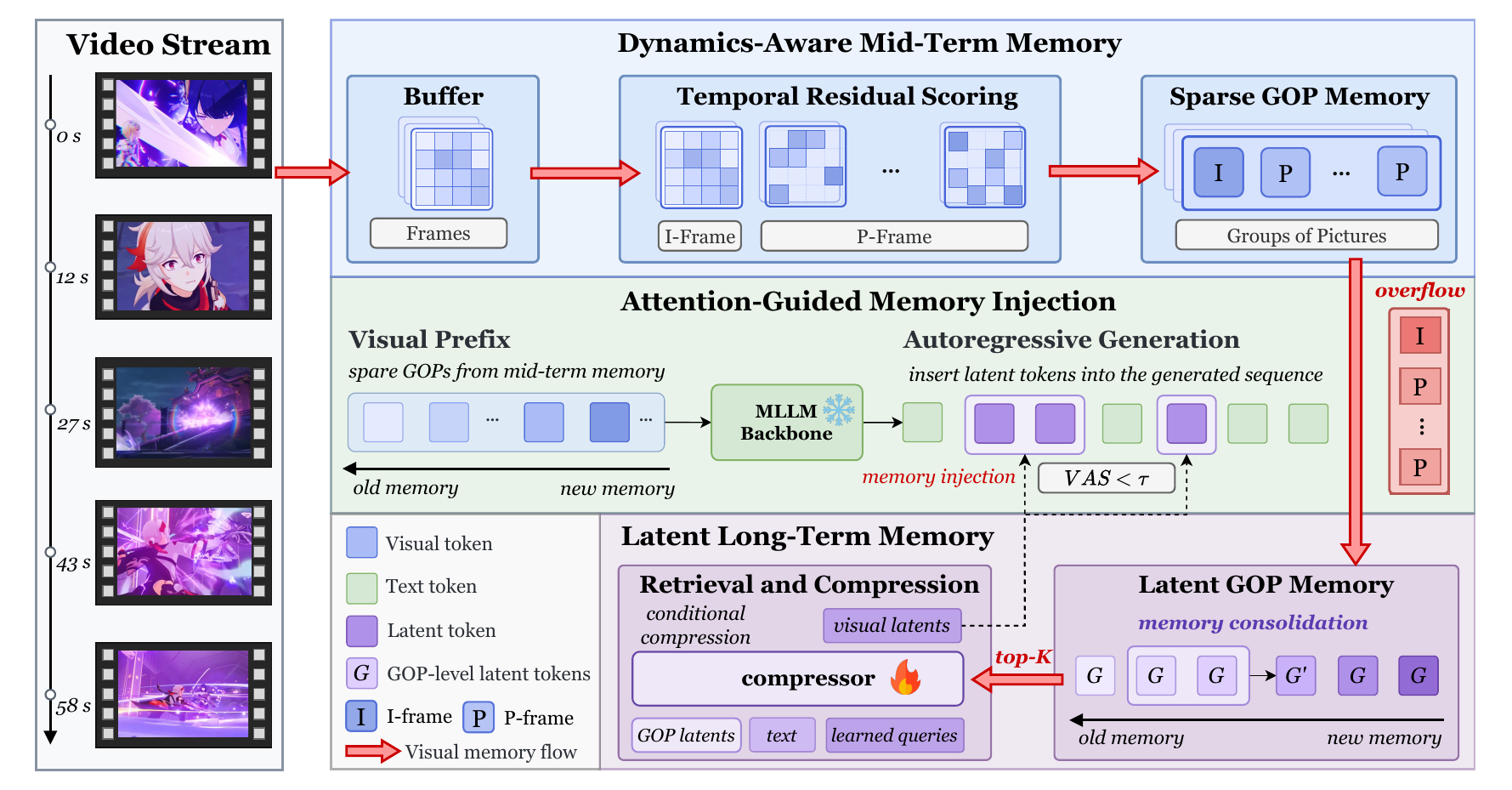}
    \caption{Overview of \ourmethod. $(1)$ Dynamics-aware mid-term memory organizes incoming frames into GOPs using temporal residual scoring. $(2)$ Latent long-term memory encodes overflowing GOPs as visual latents and consolidates similar GOPs. $(3)$ Attention-guided injection retrieves query-relevant GOPs, compresses them into visual latents, and injects them when the visual attention score falls below a threshold.}
    \vspace{-0.5em}
    \label{fig:method}
\end{figure*}

\section{Methodology}
In this section, we first present the overall pipeline of StreamFlow ($\triangleright$ \Cref{sec:pipeline}). We then describe its two memory modules in detail: the dynamics-aware mid-term memory ($\triangleright$ \Cref{sec:midterm}) and the latent long-term memory ($\triangleright$ \Cref{sec:longterm}). Finally, we introduce the attention-guided memory injection mechanism ($\triangleright$ \Cref{sec:injection}), which continuously monitors the model's reliance on visual information during generation and injects relevant visual latents on demand.

\subsection{Overall Pipeline}\label{sec:pipeline}
As illustrated in \Cref{fig:method}, \ourmethod comprises a dynamics-aware mid-term memory $\mathcal{M}_{\mathrm{M}}$ and a latent long-term memory $\mathcal{M}_{\mathrm{L}}$, which preserve visual information at different temporal scales. As video frames continuously arrive, the mid-term memory selectively retains dynamic visual patches from incoming frames before visual encoding. Once the mid-term memory is full, older visual information is encoded into GOP-level visual latents and transferred to the long-term memory for later retrieval. During answer generation, \ourmethod monitors the MLLM's attention to visual information and injects visual latents retrieved from the long-term memory when needed, enabling adaptive reference to historical visual evidence and mitigating attention drift.

\subsection{Dynamics-Aware Mid-Term Memory}\label{sec:midterm}
To support real-time efficient spatiotemporal perception, \ourmethod maintains a bounded dynamics-aware mid-term memory $\mathcal{M}_{\mathrm{M}}$. It partitions the video stream into consecutive groups of pictures (GOPs), identifies temporal changes from raw pixel differences, and selectively encodes dynamic patches. By filtering redundant content before visual encoding, it reduces unnecessary encoding computation while preserving fine-grained recent dynamics.

\textbf{Reference-Anchored Grouping.} As a basis for quantifying local temporal changes, the mid-term memory organizes the video stream into consecutive GOPs, each comprising a single I-frame and multiple P-frames. Specifically, as video frames arrive, \ourmethod temporarily stores them in their original form in a frame buffer. Once the buffer accumulates $T$ frames, they are grouped into the $g$-th GOP, denoted as $\mathcal{G}^{g}=\{\mathbf{X}^{g}_{0},\mathbf{X}^{g}_{1},\ldots,\mathbf{X}^{g}_{T-1}\}$, and passed to the subsequent patch-level temporal residual scoring module. Within this module, $\mathbf{X}^{g}_{0}$ serves as the shared I-frame reference, while the remaining frames serve as P-frames for measuring local temporal variations against the reference.

\textbf{Patch-Level Residual Scoring.} Streaming videos exhibit substantial local temporal redundancy: over short intervals, most content remains stable, with new visual evidence concentrated in a few dynamic regions. Motivated by this observation, \ourmethod uses temporal residuals within each GOP to localize these changes directly in raw pixel space before visual encoding. Given the reference frame $\mathbf{X}^{g}_{0}$, we construct a residual field for each P-frame $\mathbf{X}^{g}_{f}$ as
\begin{equation}
\mathbf{R}^{g}_{f}(\mathbf{u})
=
\mathbf{X}^{g}_{f}(\mathbf{u})
-
\mathbf{X}^{g}_{0}(\mathbf{u}),
\qquad f=1,\ldots,T-1,
\end{equation}
where $\mathbf{u}$ denotes a spatial location and $\mathbf{R}^{g}_{f}(\mathbf{u})\in\mathbb{R}^{3}$ is the corresponding RGB residual. Since scene dynamics typically produce spatially localized residual responses, we aggregate residual magnitudes within each visual patch to reduce sensitivity to pixel-level noise. For the $i$-th patch covering the pixel set $\mathcal{P}_{i}$, its temporal residual score is defined as
\begin{equation}
s^{g}_{f,i}
=
\frac{1}{|\mathcal{P}_{i}|}
\sum_{\mathbf{u}\in\mathcal{P}_{i}}
\left\|\mathbf{R}^{g}_{f}(\mathbf{u})\right\|_{2}.
\end{equation}
This reference-anchored score provides a dynamics-aware prior that identifies where new temporal evidence emerges and guides subsequent sparse patch selection.

\textbf{Sparse Patch Selection.} Guided by temporal residual scores, \ourmethod allocates asymmetric patch budgets across each GOP to suppress temporally redundant content while preserving informative dynamics. All $N$ I-frame patches are retained as the spatial reference, whereas each P-frame keeps the top $\lceil\rho N\rceil$ patches, with $\rho\in(0,1]$. Formally, the selected index set maximizes the aggregate residual score under this retention constraint:
\begin{equation}
\mathcal{S}^{g}_{f}
=
\operatorname*{arg\,max}_{\substack{
\mathcal{I}\subseteq\{1,\ldots,N\}\\
|\mathcal{I}|=\lceil \rho N\rceil}}
\sum_{i\in\mathcal{I}} s^{g}_{f,i}.
\end{equation}

The selected patches are then chronologically arranged to form a sparse GOP $\widehat{\mathcal{G}}^{g}$, which preserves the most informative visual content and is subsequently incorporated into the mid-term memory $\mathcal{M}_{\mathrm{M}}$.

\subsection{Latent Long-Term Memory}\label{sec:longterm}
To preserve historical visual information displaced from $\mathcal{M}_{\mathrm{M}}$, \ourmethod introduces a latent long-term memory $\mathcal{M}_{\mathrm{L}}$, which encodes earlier video segments as GOP-level visual latents that remain accessible to subsequent reasoning. By consolidating visually redundant GOPs through I-frame-guided temporal subsampling and token pruning, $\mathcal{M}_{\mathrm{L}}$ progressively consolidates historical evidence within a fixed memory budget as the stream evolves.

\textbf{Visual Latent Encoding.} When a sparse GOP $\widehat{\mathcal{G}}^{g}$ leaves the mid-term memory, \ourmethod encodes its retained patches at their original spatiotemporal coordinates using the frozen visual encoder $\mathcal{E}_{v}$ of the backbone MLLM:
\begin{equation}
\mathbf{H}^{g}
=\mathcal{E}_{v}\left(\widehat{\mathcal{G}}^{g}\right)
=\left[\mathbf{h}^{g}_{1}, \ldots, \mathbf{h}^{g}_{N_g}\right]^{\top}
\in \mathbb{R}^{N_g\times d_v},
\end{equation}
where $N_g$ denotes the number of retained visual tokens and $d_v$ is the hidden dimension of the visual encoder. The resulting query-independent visual latents preserve the selected spatiotemporal structure of the GOP.

\textbf{Memory Management.} Directly appending incoming visual latents would make $\mathcal{M}_{\mathrm{L}}$ grow unbounded, while FIFO eviction would discard early visual evidence. \ourmethod instead maintains a temporally ordered latent memory with a fixed capacity of $C$ GOPs:
\begin{equation}
\mathcal{M}_{\mathrm{L}}
=
\left[
\mathbf{H}^{(1)},
\ldots,
\mathbf{H}^{(n)}
\right],
\qquad 0\le n\le C,
\end{equation}
where $\mathbf{H}^{(j)}=[\mathbf{I}_{j},\mathbf{P}^{1}_{j},\ldots,\mathbf{P}^{T-1}_{j}]$ is the $j$-th GOP representation and the entries are ordered from oldest to newest. If $n<C$, the incoming GOP $\mathbf{H}^{\mathrm{new}}$ is appended directly. Once the memory reaches capacity, \ourmethod selects the most similar adjacent pair by averaging the cosine similarities between spatially corresponding tokens of their I-frames:
\begin{equation}
\begin{aligned}
s_j
&=
\frac{1}{N}
\sum_{i=1}^{N}
\operatorname{cos}
\left(
\mathbf{I}_{j,i},
\mathbf{I}_{j+1,i}
\right),\\
p
&=
\operatorname*{arg\,max}_{j\in\{1,\ldots,C-1\}}s_j,
\end{aligned}
\end{equation}
where $N$ is the number of I-frame tokens. The merge operator $\mathcal{F}_{\mathrm{merge}}$ anchors on the earlier I-frame, retains the $\lceil\rho N\rceil$ least similar tokens from the later I-frame as a sparse P-frame, and uniformly subsamples the combined $2T$ frames to $T$ frames:
\begin{equation}
\overline{\mathbf{H}}^{(p)}
=
\mathcal{F}_{\mathrm{merge}}
\left(\mathbf{H}^{(p)},\mathbf{H}^{(p+1)}\right).
\end{equation}
The merged representation replaces the selected pair, freeing one slot for the incoming GOP $\mathbf{H}^{\mathrm{new}}$.
This memory consolidation preserves temporal order while maintaining bounded access to long-range visual information.

\subsection{Attention-Guided Memory Injection}\label{sec:injection}
Prepending the entire streaming memory to the MLLM input is inefficient: it consumes substantial context, while attention to the visual prefix decays during autoregressive generation. \ourmethod instead uses only mid-term memory as the visual prefix and injects long-term latents on demand based on the visual attention score (VAS), enabling dynamic access to historical evidence with limited context overhead.

\textbf{Visual Attention Score.} To efficiently track the model's reliance on visual evidence during decoding, we use VAS as an online grounding signal, motivated by recent findings in MLLMs~\citep{luo2026narrow,wu2026spotlight}. For the $t$-th generated token, let $A_{t,c}^{\ell,h}$ denote the normalized attention weight to context token $c$ at layer $\ell$ and head $h$, and let $\mathcal{V}_{t}$ be the indices of accessible visual tokens. VAS is defined as the visual attention mass averaged across all layers and heads:
\begin{equation}
\operatorname{VAS}_{t}
=
\frac{1}{N_{\ell}N_{h}}
\sum_{\ell=1}^{N_{\ell}}
\sum_{h=1}^{N_{h}}
\sum_{c\in\mathcal{V}_{t}}
A_{t,c}^{\ell,h}.
\end{equation}
A low $\operatorname{VAS}_{t}$ indicates that linguistic context is dominating visual evidence, prompting \ourmethod to reactivate long-term memory for subsequent generation.

\textbf{Memory Retrieval.} Once the model's visual attention score falls below the threshold $\tau$, \ourmethod retrieves the $K$ GOPs most relevant to the current reasoning state. Let $\mathcal{Q}_{t}$ denote the retrieval-query tokens and $\mathbf{e}_{u}$ the input embedding of token $u\in\mathcal{Q}_{t}$. Using the backbone's frozen visual projector $\mathcal{P}_{v}$, we map visual tokens into the text embedding space and construct a pooled query and frame-level keys:
\begin{equation}
\mathbf{q}_{t}
=
\operatorname{Avg}_{u\in\mathcal{Q}_{t}}\mathbf{e}_{u},
\quad
\mathbf{k}_{j,f}
=
\operatorname{Avg}_{i=1}^{N_{j,f}}
\mathcal{P}_{v}\!\left(\mathbf{h}_{f,i}^{(j)}\right).
\end{equation}
Here, $\mathbf{h}_{f,i}^{(j)}$ denotes the $i$-th valid token of frame $f$ in GOP $j$, and $N_{j,f}$ is the number of valid tokens.
Each GOP is scored by its most relevant frame, after which the top-$K$ GOPs are restored to chronological order:
\begin{equation}
\begin{aligned}
r_{t,j}
&=
\max_{f\in\{1,\ldots,T\}}
\operatorname{cos}\!\left(\mathbf{q}_{t},\mathbf{k}_{j,f}\right),\\
\mathcal{I}_{t}
&=
\operatorname{sort}\!\left(
\operatorname{TopK}_{j\in\{1,\ldots,n\}}\!\left(r_{t,j}\right)
\right).
\end{aligned}
\end{equation}
A neural compressor $\mathcal{C}_{\phi}$ parameterized by $\phi$ then jointly processes the chronologically concatenated GOPs $\mathbf{H}^{(\mathcal{I}_{t})}$, the query tokens, and $L$ learned queries:
\begin{equation}
\mathbf{Z}_{t}
=
\mathcal{C}_{\phi}\!\left(
\mathcal{P}_{v}\!\left(\mathbf{H}^{(\mathcal{I}_{t})}\right),
\mathcal{Q}_{t};
\mathbf{L}
\right).
\end{equation}
The compressor retains only the outputs of the $L$ learned queries as a fixed visual bottleneck, and the resulting latents $\mathbf{Z}_{t}$ are inserted after the current decoding prefix, thereby enabling dynamic access to relevant long-term visual evidence during subsequent token generation.

\section{Experiments}

\subsection{Experimental Setup}

\textbf{Benchmarks.}
We evaluate \ourmethod on both offline and streaming video benchmarks. For offline evaluation, we adopt MVBench~\citep{li2024mvbench}, MLVU~\citep{zhou2025mlvu}, and VideoMME~\citep{fu2025video}, while for streaming evaluation, we use StreamingBench~\citep{lin2026streamingbench}. Across all benchmarks, we adopt a unified streaming evaluation protocol, where videos are presented frame by frame and each question is revealed only after its associated end timestamp, requiring causal reasoning over previously observed content~\citep{chen2026streamingtom}. Further details of the benchmarks are provided in Appendix~\ref{app:benchmark_descriptions}.

\textbf{Baselines.}
We compare \ourmethod against two categories of approaches to streaming video understanding: model-based and memory-based methods. The model-based baselines include VideoLLM-Online~\citep{chen2024videollm}, Dispider~\citep{qian2025dispider}, TimeChat-Online~\citep{yao2025timechat}, and StreamForest~\citep{zeng2026streamforest}. The memory-based baselines include LiveVLM~\citep{ning2025livevlm}, StreamMem~\citep{yang2025streammem}, InfiniPot-V~\citep{kim2026infinipot}, StreamingTOM~\citep{chen2026streamingtom}, HERMES~\citep{zhang2026hermes}, FluxMem~\citep{xie2026fluxmem}, and CausalMem~\citep{song2026towards}. Further details of the baselines and their configurations are provided in Appendix~\ref{app:baseline_setup}.

\textbf{Training Data.}
The training data consist of question-conditioned video examples drawn from COIN~\citep{tang2019coin} for multi-step procedures, NeXT-QA~\citep{xiao2021next} for temporally dependent event reasoning, STAR~\citep{wu2024star} for interactions and action transitions, CLEVRER~\citep{yi2019clevrer} for controlled object dynamics, and selected LLaVA-Video-178K data~\citep{zhang2024llava} for broader real-world coverage. We use Qwen3.5-27B to construct the supervision labels for these training samples. Dataset details are provided in Appendices~\ref{app:training_data} and~\ref{app:implementation_details}.

\textbf{Implementation Details.}
We use Qwen3.5-9B as the frozen MLLM backbone and adapt the same model with LoRA as the memory compressor. The mid-term memory accommodates up to $32$ frames, organized into four-frame GOPs, and the P-frame retention ratio $\rho$ is set to $0.5$. The long-term memory maintains up to $96$ GOPs through similarity-guided consolidation. At each memory injection, we retrieve $K=4$ GOPs and compress them into $L=32$ visual latents. We set the VAS threshold to $\tau=0.1$. The compressor is trained via supervised fine-tuning (SFT) on four NVIDIA L20X GPUs, while the answering backbone remains frozen. During evaluation, videos are sampled at $1$ FPS and processed in a streaming manner. See Appendix~\ref{app:implementation_details} for implementation details.

\begin{table*}[t]
    \centering
    \scriptsize
    \setlength{\tabcolsep}{0.3pt}
    \renewcommand{\arraystretch}{1.0}
    \definecolor{tblgroup}{gray}{0.95}
    \definecolor{tblours}{gray}{0.88}
    \caption{
    Performance comparison on video understanding benchmarks. The \#Frames column reports either the number of input frames or the sampling rate (fps), following each method's original evaluation setting. ``--'' denotes unreported results. The best and second-best results among all compared methods are in \textbf{bold} and \underline{underlined}, respectively.
    }
    \label{tab:main_results}
    \begin{tabularx}{\textwidth}{@{}l c@{\hspace{3pt}}|@{\hspace{3pt}}>{\hsize=1.45\hsize\centering\arraybackslash}X >{\hsize=1.05\hsize\centering\arraybackslash}X >{\hsize=1.00\hsize\centering\arraybackslash}X *{3}{>{\hsize=0.80\hsize\centering\arraybackslash}X} >{\hsize=1.10\hsize\centering\arraybackslash}X@{}}
    \toprule
    \multirow{2}{*}{\textbf{Method}} &
    \multirow{2}{*}{\textbf{\#Frames}} &
    \multicolumn{1}{c}{\textbf{StreamingBench}} &
    \multirow{2}{*}{\textbf{MVBench}} &
    \multirow{2}{*}{\textbf{MLVU}} &
    \multicolumn{4}{c}{\textbf{VideoMME}} \\
    \cmidrule(lr){3-3}\cmidrule(l){6-9}
    & & \textbf{Overall} & & & \textbf{S} & \textbf{M} & \textbf{L} & \textbf{Overall} \\
    \midrule
    \rowcolor{tblgroup}
    \multicolumn{9}{@{}c@{}}{\textbf{Model-based}} \\
    VideoLLM-Online-8B~\cite{chen2024videollm} & 2 fps & 32.48 & -- & -- & -- & -- & -- & -- \\
    Dispider-7B~\cite{qian2025dispider} & 1 fps & 53.12 & -- & 61.70 & -- & 53.70 & 49.70 & 57.20 \\
    TimeChat-Online-7B~\cite{yao2025timechat} & 1 fps & 58.11 & -- & 65.40 & -- & -- & 52.40 & 63.30 \\
    StreamForest-7B~\cite{zeng2026streamforest} & 1 fps & -- & \textbf{70.20} & 70.00 & -- & -- & -- & 61.40 \\
    \addlinespace[1.5pt]
    \rowcolor{tblgroup}
    \multicolumn{9}{@{}c@{}}{\textbf{Memory-based}} \\
    LiveVLM-7B~\cite{ning2025livevlm} & 0.5/0.2 fps & \underline{63.10} & -- & 68.10 & -- & 57.00 & 51.30 & 59.60 \\
    StreamMem-7B~\cite{yang2025streammem} & 0.5/0.2 fps & -- & -- & 66.90 & -- & 56.60 & 50.10 & 59.40 \\
    InfiniPot-V-7B~\cite{kim2026infinipot} & 0.5 fps & -- & -- & 65.80 & 74.10 & 60.80 & 53.40 & 62.80 \\
    StreamingTOM-7B~\cite{chen2026streamingtom} & 0.5 fps & -- & -- & 67.90 & 71.30 & 57.80 & 50.60 & 59.90 \\
    HERMES-7B~\cite{zhang2026hermes} & 1 fps & -- & 65.53 & -- & -- & -- & 53.44 & 60.63 \\
    FluxMem-7B~\cite{xie2026fluxmem} & 1 fps & -- & -- & \underline{73.10} & \underline{76.90} & \underline{65.10} & \underline{54.00} & \underline{65.30} \\
    CausalMem-7B~\cite{song2026towards} & 0.5/0.2 fps & -- & -- & 70.90 & -- & -- & -- & 60.00 \\
    \rowcolor{tblours}[3.5pt][3.5pt]
    \textbf{\ourmethod{}-9B (ours)} & \multicolumn{1}{c@{\hspace{\dimexpr6pt+\arrayrulewidth\relax}}}{1 fps} & \textbf{67.73} & \underline{68.90} & \textbf{75.34} & \textbf{87.78} & \textbf{70.67} & \textbf{62.11} & \textbf{73.52} \\
    \bottomrule
    \end{tabularx}
\end{table*}

\begin{table*}[t]
    \centering
    \scriptsize
    \setlength{\tabcolsep}{1.0pt}
    \renewcommand{\arraystretch}{1.0}
    \definecolor{tblgroup}{gray}{0.95}
    \definecolor{tblours}{gray}{0.88}
    \caption{
    Accuracy (\%) on all ten task types in the Real-Time Visual Understanding (RTVU) subset of StreamingBench. ``--'' denotes an unreported result. The best and second-best results among all methods are in \textbf{bold} and \underline{underlined}, respectively.
    }
    \label{tab:streamingbench_rtvu_breakdown}
    \begin{tabularx}{\textwidth}{@{}l *{11}{>{\centering\arraybackslash}X}@{}}
    \toprule
    \textbf{Method} &
    \textbf{OP} &
    \textbf{CR} &
    \textbf{CS} &
    \textbf{ATP} &
    \textbf{EU} &
    \textbf{TR} &
    \textbf{PR} &
    \textbf{SU} &
    \textbf{ACP} &
    \textbf{CT} &
    \textbf{Overall} \\
    \midrule
    \rowcolor{tblgroup}
    \multicolumn{12}{@{}c@{}}{\textbf{Model-based}} \\
    VideoLLM-Online-8B~\cite{chen2024videollm} & 39.07 & 40.06 & 34.49 & 31.05 & 45.96 & 32.40 & 31.48 & 34.16 & 42.49 & 27.89 & 35.99 \\
    Dispider-7B~\cite{qian2025dispider} & 74.92 & 75.53 & 74.10 & 73.08 & 74.44 & 59.92 & 76.14 & 62.91 & 62.16 & 45.80 & 67.63 \\
    TimeChat-Online-7B~\cite{yao2025timechat} & 80.22 & 82.03 & 79.50 & 83.33 & 76.10 & 78.50 & 78.70 & 64.63 & 69.60 & \underline{57.98} & 75.36 \\
    StreamForest-7B~\cite{zeng2026streamforest} & 83.11 & \underline{82.81} & 82.65 & 84.26 & 77.50 & 78.19 & 76.85 & 69.11 & \underline{75.64} & 54.40 & 77.26 \\
    \addlinespace[1.5pt]
    \rowcolor{tblgroup}
    \multicolumn{12}{@{}c@{}}{\textbf{Memory-based}} \\
    LiveVLM-7B~\cite{ning2025livevlm} & 79.84 & 79.69 & 84.86 & 80.72 & 67.08 & 70.09 & 74.07 & 66.26 & 67.99 & 42.49 & 72.36 \\
    InfiniPot-V-7B~\cite{kim2026infinipot} & -- & -- & -- & -- & -- & -- & -- & -- & -- & -- & 76.40 \\
    HERMES-7B~\cite{zhang2026hermes} & \underline{83.65} & 81.25 & \textbf{88.01} & \textbf{87.46} & 76.73 & \textbf{86.60} & \underline{82.41} & \underline{76.02} & 73.94 & 46.63 & \underline{79.44} \\
    FluxMem-7B~\cite{xie2026fluxmem} & 80.20 & 81.10 & 81.40 & 85.30 & \underline{78.00} & \underline{83.80} & 80.60 & 65.90 & 69.60 & 52.10 & 76.40 \\
    CausalMem-7B~\cite{song2026towards} & 82.60 & 79.70 & 83.20 & 83.20 & 69.60 & 78.40 & 75.00 & 67.30 & 70.90 & 39.40 & 74.30 \\
    \rowcolor{tblours}
    \textbf{\ourmethod{}-9B (ours)} & \textbf{84.01} & \textbf{87.50} & \underline{87.38} & \underline{85.53} & \textbf{82.28} & 80.69 & \textbf{88.89} & \textbf{78.05} & \textbf{76.42} & \textbf{67.02} & \textbf{81.55} \\
    \bottomrule
    \end{tabularx}
\end{table*}

\subsection{Main Results}

\textbf{\ourmethod Achieves Consistent Improvements Across Streaming Video Understanding Tasks.}
As shown in \Cref{tab:main_results}, \ourmethod achieves an overall score of $67.73\%$ on StreamingBench, surpassing the strongest baseline, LiveVLM-7B, by $4.63\%$. The RTVU breakdown in \Cref{tab:streamingbench_rtvu_breakdown} further shows that this improvement is consistent across task types: \ourmethod attains an overall score of $81.55\%$, exceeding HERMES-7B by $2.11\%$, and ranks first on seven of the ten task types. Most notably, \ourmethod achieves $67.02\%$ on Counting (CT), outperforming TimeChat-Online-7B by $9.04\%$. Moreover, for a fairer comparison, we reimplemented selected baselines with the same Qwen3.5-9B backbone and otherwise identical settings. As reported in Appendix~\ref{app:additional_benchmark_results}, \ourmethod outperforms the strongest baseline, StreamingTOM, by $1.17\%$ on StreamingBench. These results demonstrate the effectiveness of the \ourmethod across diverse streaming video understanding tasks.

\textbf{\ourmethod Achieves Strong Offline Video Understanding.}
As shown in \Cref{tab:main_results}, \ourmethod demonstrates consistently strong performance on both MLVU and VideoMME. On MLVU, \ourmethod obtains $75.34\%$, outperforming the strongest baseline, FluxMem-7B, by $2.24\%$. Moreover, on VideoMME, \ourmethod achieves the best results across the short, medium, and long subsets, with scores of $87.78\%$, $70.67\%$, and $62.11\%$, respectively, leading to an overall score of $73.52\%$. These results exceed the corresponding second-best scores by $10.88\%$, $5.57\%$, $8.11\%$, and $8.22\%$. Under the same Qwen3.5-9B backbone and otherwise identical settings, \ourmethod also outperforms the strongest baseline, ReKV, by $1.63\%$ on VideoMME, as reported in Appendix~\ref{app:additional_benchmark_results}. The consistent improvements across different video durations suggest that \ourmethod preserves both recent visual details and long-range evidence while enabling the frozen MLLM backbone to remain effective on conventional offline video understanding.

\Needspace{18\baselineskip}
\subsection{Framework Analysis}

\rightwrapfiguretight{%
\vspace{-0.5\baselineskip}
\includegraphics[width=\linewidth,trim=0 2bp 0 0,clip]{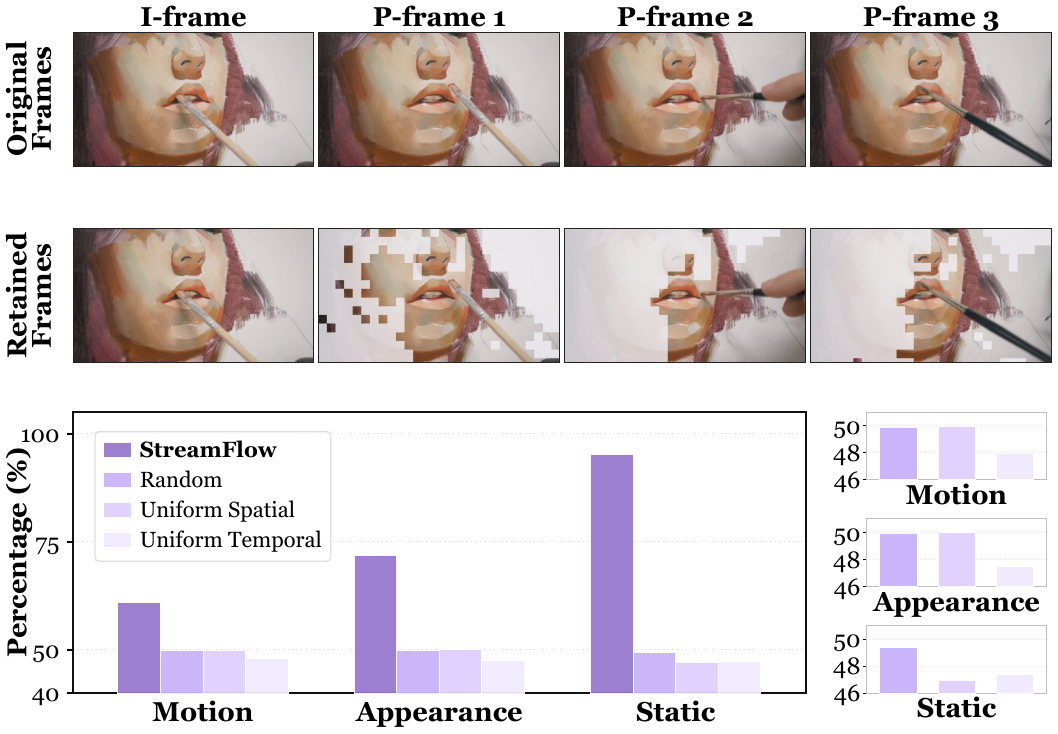}
}{Patch-level residual scoring preserves dynamic visual content while rejecting static patches.}{fig:patch_residual}
\textbf{Patch-Level Residual Scoring Preserves Dynamic Visual Content.} \Cref{fig:patch_residual} compares four patch-retention policies under $50\%$ P-frame retention and an identical GOP-wide budget (see Appendix~\ref{app:retention_ratio_masks} for baseline details). By concentrating on dynamic regions rather than repeated background, \ourmethod preserves $61.08\%$ of camera-motion-compensated local motion, versus $49.93\%$ for the strongest alternative, while rejecting $95.28\%$ of independently identified static patches, versus at most $49.37\%$. It similarly retains $71.96\%$ of motion-compensated appearance change, compared with $50.03\%$ for the strongest alternative. These results show that patch-level residual scoring prioritizes evolving visual evidence without repeatedly allocating memory to static regions already captured by the complete I-frame.
\par\WFclear

\textbf{\ourmethod Sustains Visual Grounding.} \Cref{fig:attention_injection}(a) reveals attention drift: as the linguistic context grows, all visual-prefix baselines attend less to visual evidence. In contrast, \ourmethod maintains higher visual attention throughout decoding, raising mean VAS from $0.066$ under Vanilla to $0.105$, a relative gain of $59.1\%$. We next examine whether this restoration of visual attention is accompanied by improved grounding in relevant historical evidence. In \Cref{fig:attention_injection}(b), the shuffled condition replaces each sample's matched long-term memory with memory from another sample, while all other inputs, model parameters, and the visual-token budget remain fixed.
\inlinewrapfigure{%
\includegraphics[width=\linewidth,trim=0 16bp 0 0,clip]{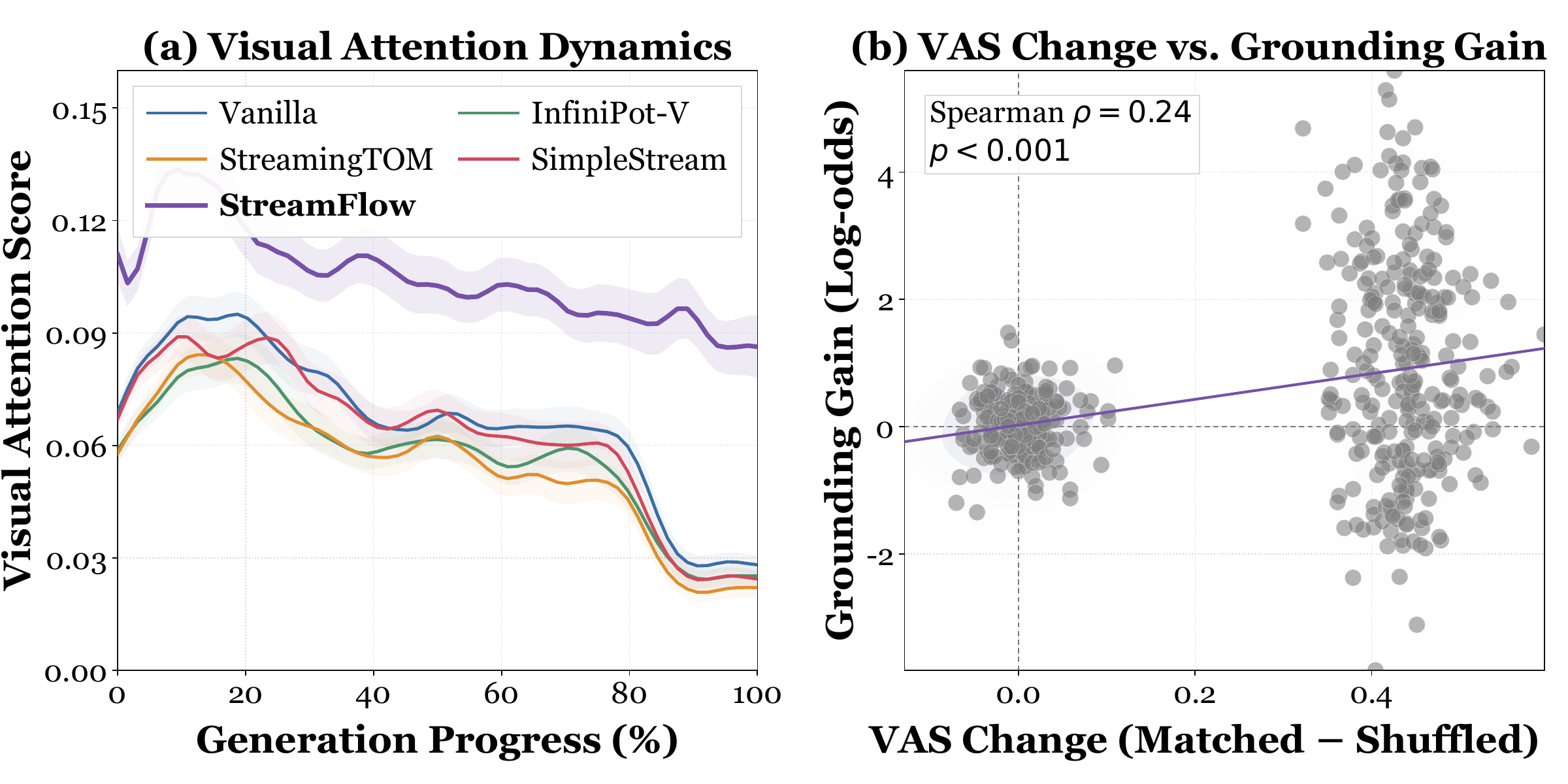}
}{Generation-time VAS and grounding gain under matched versus shuffled memory.}{fig:attention_injection}
Compared with shuffled memory, matched memory increases correct-answer log-odds by $0.27$ on average, with larger VAS gains associated with greater grounding gains (Spearman $\rho=0.24$, $p<0.001$). By changing only memory relevance, this intervention shows that \ourmethod's higher VAS is tied to the use of relevant historical evidence. Together, these results demonstrate that \ourmethod sustains visual grounding throughout reasoning. Details of the VAS computation and counterfactual intervention are provided in Appendices~\ref{app:vas_computation} and~\ref{app:counterfactual_intervention}; case studies in Appendix~\ref{app:long_video_cases} show how VAS-guided insertion restores long-range evidence.
\par\WFclear

\rightwraptable{%
\centering
\small
\setlength{\tabcolsep}{1.8pt}
\definecolor{tblgroup}{gray}{0.95}
\begin{tabular*}{\linewidth}{@{\extracolsep{\fill}}lccc@{}}
    \toprule
    Metric $\downarrow$ & Vanilla & \ourmethod & Red. \\
    \midrule
    \rowcolor{tblgroup}
    \multicolumn{4}{c}{\textbf{Visual front end}} \\
    Pre-ViT patches & 36,318 & \textbf{22,781} & 37.3\% \\
    VFE latency (ms) & 450.4 & \textbf{225.1} & 50.0\% \\
    \addlinespace[1.5pt]
    \rowcolor{tblgroup}
    \multicolumn{4}{c}{\textbf{Decoder}} \\
    Context length & 9,161 & \textbf{5,700} & 37.8\% \\
    KV-cache memory (MiB) & 1,288.3 & \textbf{801.5} & 37.8\% \\
    Attention latency (ms) & 3,162.6 & \textbf{1,156.4} & 63.4\% \\
    \addlinespace[1.5pt]
    \rowcolor{tblgroup}
    \multicolumn{4}{c}{\textbf{Complete task}} \\
    Task peak memory (GiB) & 13.68 & \textbf{10.80} & 21.1\% \\
    Task latency (s) & 7.36 & \textbf{3.65} & 50.4\% \\
    \bottomrule
\end{tabular*}

}{Inference efficiency on RTVU across the visual front end, decoder, and complete task. Stage-level rows isolate their named scope, while complete-task rows cover the full inference pipeline.}{tab:rtvu_efficiency}
\textbf{Pre-ViT Sparsification and Bounded Memory Reduce Inference Overhead.} \Cref{tab:rtvu_efficiency} reports efficiency at three levels: the visual front end, the decoder, and the complete task (see Appendix~\ref{app:efficiency_boundaries}). At the visual front end, Vanilla encodes every patch, whereas \ourmethod retains the complete I-frame and only the highest-residual $50\%$ of P-frame patches within each GOP. On RTVU, this selective encoding reduces the mean Pre-ViT patch count by $37.3\%$. Crucially, even after accounting for residual scoring and sparse packing, visual-front-end latency decreases by $50.0\%$, confirming that patch selection yields substantial net savings. At the decoder, bounded memory shortens the context from $9{,}161$ to $5{,}700$ tokens ($37.8\%$), correspondingly reduces KV-cache memory by $37.8\%$, and cuts self-attention latency by $63.4\%$. These stage-level gains reduce end-to-end latency by $50.4\%$ and peak memory by $21.1\%$.

\subsection{Sensitivity Analysis and Ablation Studies}

\rightwrapfigure{%
\vspace{-0.8\baselineskip}
\includegraphics[width=\linewidth,trim=0 3bp 0 0,clip]{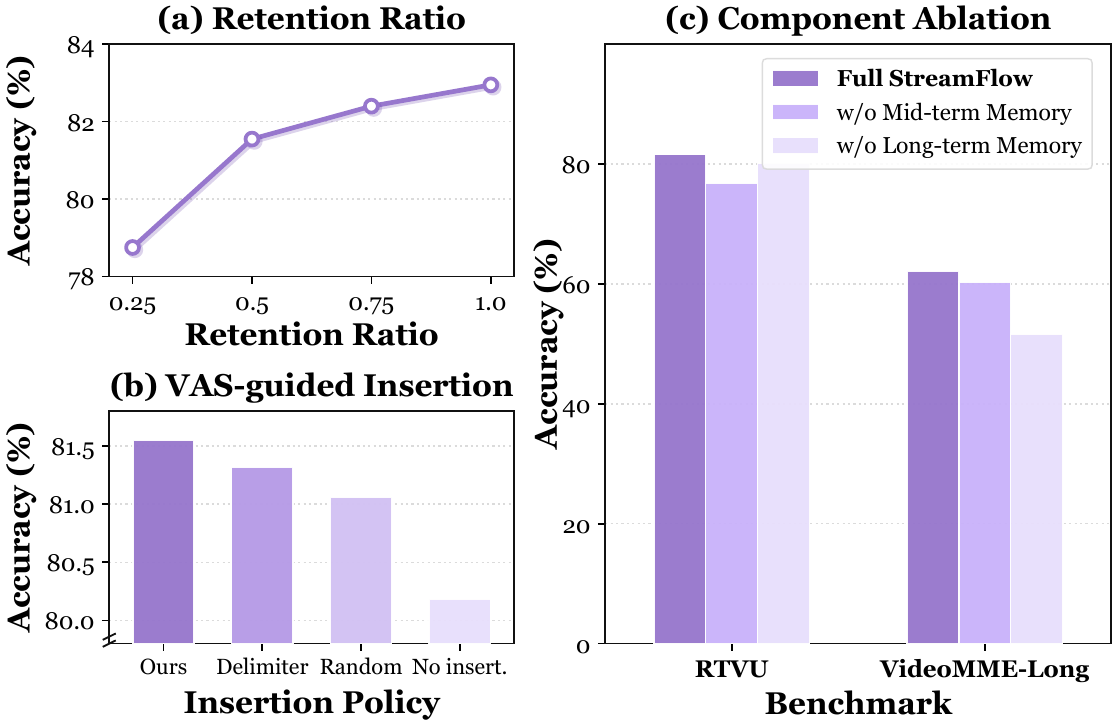}
}{(a) Retention-ratio sensitivity on RTVU. (b) Insertion-policy sensitivity on RTVU. (c) Component ablation on RTVU and VideoMME-Long.}{fig:sensitivity_ablation}
\textbf{Sensitivity Analysis.} On RTVU, \Cref{fig:sensitivity_ablation}(a) analyzes the retention ratio $\rho$. Increasing the ratio from $0.25$ to $0.50$ improves accuracy from $78.75\%$ to $81.55\%$, while subsequent increases to $0.75$ and $1.00$ yield smaller gains of $0.85\%$ and $0.55\%$. This clear diminishing-return trend indicates that retaining half of the P-frame patches ranked by residual score captures most of the useful intermediate evidence, motivating the default ratio of $0.50$ as a balance between accuracy and visual-token cost. \Cref{fig:sensitivity_ablation}(b) further examines the generation-time insertion policy. VAS-guided insertion achieves $81.55\%$ accuracy, outperforming budget-matched delimiter and random insertion by $0.23\%$ and $0.49\%$, respectively, indicating that VAS provides a reliable signal for timely memory access. Disabling insertion, which effectively removes access to long-term memory, reduces accuracy by $1.37\%$ to $80.18\%$. These results suggest that the learned visual latents are relatively robust to their exact insertion positions, while VAS-guided insertion provides a consistent additional benefit by determining when the memory should be activated during subsequent generation. Additional analyses are provided in Appendix~\ref{app:sensitivity_analysis}.

\textbf{Component Ablation.}
\Cref{fig:sensitivity_ablation}(c) isolates the contribution of the two memory timescales on RTVU and VideoMME-Long. All variants share the same evaluation protocol: removing mid-term memory drops its records from the raw visual prefix, whereas removing long-term memory disables its construction and dynamic insertion. Removing mid-term memory reduces accuracy from $81.55\%$ to $76.86\%$ on RTVU and from $62.11\%$ to $60.33\%$ on VideoMME-Long, corresponding to drops of $4.69\%$ and $1.78\%$, respectively. Removing long-term memory lowers accuracy to $80.18\%$ on RTVU and $51.67\%$ on VideoMME-Long, corresponding to drops of $1.37\%$ and $10.44\%$. Thus, high-fidelity mid-term evidence contributes more strongly to real-time understanding, whereas compressed long-term memory is critical for the longest offline videos. Together, these results establish the complementarity of the two pathways: mid-term memory supplies detailed recent context, while long-term memory preserves evidence required for long-horizon reasoning.

\section{Conclusion}
In this work, we introduced \ourmethod, a memory framework that equips frozen MLLMs with efficient streaming video understanding. By combining a dynamics-aware mid-term memory with a latent long-term memory, \ourmethod selectively preserves informative visual evidence from continuously evolving video streams under a fixed memory budget. We further developed a VAS-guided memory injection mechanism that reactivates relevant long-term latents when the model's reliance on visual evidence weakens during generation, thereby mitigating visual attention drift. Extensive experiments establish state-of-the-art performance on streaming benchmarks and strong results on offline long-video understanding. Mechanistic analyses demonstrate improved visual grounding and substantial efficiency gains, validating dynamic visual-information management across encoding, consolidation, and generation.

\clearpage
\bibliographystyle{plainnat}
\bibliography{references}

@article{song2024moviechat,
  title={Moviechat: From dense token to sparse memory for long video understanding},
  author={Song, Enxin and Chai, Wenhao and Wang, Guanhong and Zhang, Yucheng and Zhou, Haoyang and Wu, Feiyang and Chi, Haozhe and Guo, Xun and Ye, Tian and Zhang, Yanting and others},
  journal={arXiv preprint arXiv:2307.16449},
  year={2023}
}

@inproceedings{qian2025dispider,
  title={Dispider: Enabling video llms with active real-time interaction via disentangled perception, decision, and reaction},
  author={Qian, Rui and Ding, Shuangrui and Dong, Xiaoyi and Zhang, Pan and Zang, Yuhang and Cao, Yuhang and Lin, Dahua and Wang, Jiaqi},
  booktitle={Proceedings of the Computer Vision and Pattern Recognition Conference},
  pages={24045--24055},
  year={2025}
}

@article{zeng2026streamforest,
  title={Streamforest: Efficient online video understanding with persistent event memory},
  author={Zeng, Xiangyu and Qiu, Kefan and Zhang, Qingyu and Li, Xinhao and Wang, Jing and Li, Jiaxin and Yan, Ziang and Tian, Kun and Tian, Meng and Zhao, Xinhai and others},
  journal={Advances in Neural Information Processing Systems},
  volume={38},
  pages={75804--75835},
  year={2026}
}

@article{guan2026video,
  title={Video streaming thinking: Videollms can watch and think simultaneously},
  author={Guan, Yiran and Yin, Liang and Liang, Dingkang and Ju, Jianzhong and Luo, Zhenbo and Luan, Jian and Liu, Yuliang and Bai, Xiang},
  journal={arXiv preprint arXiv:2603.12262},
  year={2026}
}

@article{liu2026thinking,
  title={Thinking in streaming video},
  author={Liu, Zikang and Guo, Longteng and Li, Handong and Zhen, Ru and He, Xingjian and Ji, Ruyi and Ren, Xiaoming and Zhang, Yanhao and Lu, Haonan and Liu, Jing},
  journal={arXiv preprint arXiv:2603.12938},
  year={2026}
}

@article{kim2026infinipot,
  title={Infinipot-v: Memory-constrained kv cache compression for streaming video understanding},
  author={Kim, Minsoo and Shim, Kyuhong and Choi, Jungwook and Chang, Simyung},
  journal={Advances in Neural Information Processing Systems},
  volume={38},
  pages={138983--139013},
  year={2026}
}

@inproceedings{di2025streaming,
  title={Streaming video question-answering with in-context video kv-cache retrieval},
  author={Di, Shangzhe and Yu, Zhelun and Zhang, Guanghao and Li, Haoyuan and Cheng, Hao and Li, Bolin and He, Wanggui and Shu, Fangxun and Jiang, Hao},
  booktitle={International Conference on Learning Representations},
  volume={2025},
  pages={42115--42127},
  year={2025}
}

@article{yang2025streammem,
  title={Streammem: Query-agnostic kv cache memory for streaming video understanding},
  author={Yang, Yanlai and Zhao, Zhuokai and Shukla, Satya Narayan and Singh, Aashu and Mishra, Shlok Kumar and Zhang, Lizhu and Ren, Mengye},
  journal={arXiv preprint arXiv:2508.15717},
  year={2025}
}

@article{chen2026streamingtom,
  title={Streamingtom: Streaming token compression for efficient video understanding},
  author={Chen, Xueyi and Tao, Keda and Shao, Kele and Wang, Huan},
  journal={arXiv preprint arXiv:2510.18269},
  year={2025}
}

@article{ning2025livevlm,
  title={Livevlm: Efficient online video understanding via streaming-oriented kv cache and retrieval},
  author={Ning, Zhenyu and Liu, Guangda and Jin, Qihao and Li, Chengwei and Ding, Wenchao and Guo, Minyi and Zhao, Jieru},
  journal={arXiv preprint arXiv:2505.15269},
  year={2025}
}

@inproceedings{chen2026streamkv,
  title={Streamkv: Streaming video question-answering with segment-based kv cache retrieval and compression},
  author={Chen, Yilong and Bai, Xiang and Wang, Zhibin and Bai, Chengyu and Dai, Yuhan and Lu, Ming},
  booktitle={Proceedings of the AAAI Conference on Artificial Intelligence},
  volume={40},
  number={4},
  pages={3120--3128},
  year={2026}
}

@article{shen2026simple,
  title={A simple baseline for streaming video understanding},
  author={Shen, Yujiao and Tian, Shulin and Yang, Jingkang and Liu, Ziwei},
  journal={arXiv preprint arXiv:2604.02317},
  year={2026}
}

@inproceedings{chen2024videollm,
  title={Videollm-online: Online video large language model for streaming video},
  author={Chen, Joya and Lv, Zhaoyang and Wu, Shiwei and Lin, Kevin Qinghong and Song, Chenan and Gao, Difei and Liu, Jia-Wei and Gao, Ziteng and Mao, Dongxing and Shou, Mike Zheng},
  booktitle={Proceedings of the IEEE/CVF Conference on Computer Vision and Pattern Recognition},
  pages={18407--18418},
  year={2024}
}

@inproceedings{lin2026streamingbench,
  title={Streamingbench: Assessing the gap for mllms to achieve streaming video understanding},
  author={Lin, Junming and Fang, Zheng and Chen, Chi and Cheng, Haoxuan and Wan, Zihao and Luo, Fuwen and Wang, Ziyue and Li, Peng and Liu, Yang and Sun, Maosong},
  booktitle={ICASSP 2026-2026 IEEE International Conference on Acoustics, Speech and Signal Processing (ICASSP)},
  pages={12147--12151},
  year={2026},
  organization={IEEE}
}

@article{wu2024videollm,
  title={Videollm-mod: Efficient video-language streaming with mixture-of-depths vision computation},
  author={Wu, Shiwei and Chen, Joya and Lin, Kevin Qinghong and Wang, Qimeng and Gao, Yan and Xu, Qianli and Xu, Tong and Hu, Yao and Chen, Enhong and Shou, Mike Zheng},
  journal={Advances in Neural Information Processing Systems},
  volume={37},
  pages={109922--109947},
  year={2024}
}

@inproceedings{huang2025online,
  title={Online video understanding: Ovbench and videochat-online},
  author={Huang, Zhenpeng and Li, Xinhao and Li, Jiaqi and Wang, Jing and Zeng, Xiangyu and Liang, Cheng and Wu, Tao and Chen, Xi and Li, Liang and Wang, Limin},
  booktitle={Proceedings of the Computer Vision and Pattern Recognition Conference},
  pages={3328--3338},
  year={2025}
}

@inproceedings{he2024ma,
  title={Ma-lmm: Memory-augmented large multimodal model for long-term video understanding},
  author={He, Bo and Li, Hengduo and Jang, Young Kyun and Jia, Menglin and Cao, Xuefei and Shah, Ashish and Shrivastava, Abhinav and Lim, Ser-Nam},
  booktitle={Proceedings of the IEEE/CVF conference on computer vision and pattern recognition},
  pages={13504--13514},
  year={2024}
}

@inproceedings{wang2025videollamb,
  title={Videollamb: Long streaming video understanding with recurrent memory bridges},
  author={Wang, Yuxuan and Song, Yiqi and Xie, Cihang and Liu, Yang and Zheng, Zilong},
  booktitle={Proceedings of the IEEE/CVF International Conference on Computer Vision},
  pages={24170--24181},
  year={2025}
}

@article{xie2026fluxmem,
  title={Fluxmem: Adaptive hierarchical memory for streaming video understanding},
  author={Xie, Yiweng and He, Bo and Wang, Junke and Zheng, Xiangyu and Ye, Ziyi and Wu, Zuxuan},
  journal={arXiv preprint arXiv:2603.02096},
  year={2026}
}

@article{song2026towards,
  title={Towards a Dynamic and Fixed-budget Memory Bank for Efficient Streaming Video Understanding},
  author={Song, Baiyang and Lin, Yuli and Wu, Qiong and Chen, Tao and Peng, Jun and Chen, Xiao and Zhou, Yiyi and Ji, Rongrong},
  journal={arXiv preprint arXiv:2606.25658},
  year={2026}
}

@article{chen2026scaling,
  title={Scaling the long video understanding of multimodal large language models via visual memory mechanism},
  author={Chen, Tao and Zhang, Kun and Wu, Qiong and Chen, Xiao and Chang, Chao and Sun, Xiaoshuai and Zhou, Yiyi and Ji, Rongrong},
  journal={arXiv preprint arXiv:2603.29252},
  year={2026}
}

@inproceedings{liang2026oasis,
  title={OASIS: On-Demand Hierarchical Event Memory for Streaming Video Reasoning},
  author={Liang, Zhijia and Li, Jiaming and Chen, Weikai and Zhang, Yanhao and Lu, Haonan and Li, Guanbin},
  booktitle={Proceedings of the IEEE/CVF Conference on Computer Vision and Pattern Recognition},
  pages={2821--2831},
  year={2026}
}

@article{tang2025video,
  title={Video understanding with large language models: A survey},
  author={Tang, Yunlong and Bi, Jing and Xu, Siting and Song, Luchuan and Liang, Susan and Wang, Teng and Zhang, Daoan and An, Jie and Lin, Jingyang and Zhu, Rongyi and others},
  journal={IEEE Transactions on Circuits and Systems for Video Technology},
  year={2025},
  publisher={IEEE}
}

@article{meng2026watch,
  title={Watch, Remember, Reason: Human-View Video Understanding with MLLMs},
  author={Meng, Jiahao and Tan, Yue and Xu, Qi and Gao, Kuan and Liu, Weisong and Li, Yanwei and Li, Jason and Kong, Lingdong and Wang, Haochen and Zhou, Qianyu and others},
  journal={arXiv preprint arXiv:2606.07433},
  year={2026}
}

@article{zou2024seconds,
  title={From seconds to hours: Reviewing multimodal large language models on comprehensive long video understanding},
  author={Zou, Heqing and Luo, Tianze and Xie, Guiyang and Lv, Fengmao and Wang, Guangcong and Chen, Junyang and Wang, Zhuochen and Zhang, Hansheng and Zhang, Huaijian and others},
  journal={arXiv preprint arXiv:2409.18938},
  year={2024}
}

@article{zhang2025memory,
  title={Memory in large language models: Mechanisms, evaluation and evolution},
  author={Zhang, Dianxing and Li, Wendong and Song, Kani and Lu, Jiaye and Li, Gang and Yang, Liuchun and Li, Sheng},
  journal={arXiv preprint arXiv:2509.18868},
  year={2025}
}

@article{shan2025cognitive,
  title={Cognitive memory in large language models},
  author={Shan, Lianlei and Luo, Shixian and Zhu, Zezhou and Yuan, Yu and Wu, Yong},
  journal={arXiv preprint arXiv:2504.02441},
  year={2025}
}

@inproceedings{tang2019coin,
  title={Coin: A large-scale dataset for comprehensive instructional video analysis},
  author={Tang, Yansong and Ding, Dajun and Rao, Yongming and Zheng, Yu and Zhang, Danyang and Zhao, Lili and Lu, Jiwen and Zhou, Jie},
  booktitle={Proceedings of the IEEE/CVF Conference on Computer Vision and Pattern Recognition},
  pages={1207--1216},
  year={2019}
}

@article{zhang2024llava,
  title={Llava-video: Video instruction tuning with synthetic data},
  author={Zhang, Yuanhan and Wu, Jinming and Li, Wei and Li, Bo and Ma, Zejun and Liu, Ziwei and Li, Chunyuan},
  journal={arXiv preprint arXiv:2410.02713},
  year={2024}
}

@article{wu2024star,
  title={Star: A benchmark for situated reasoning in real-world videos},
  author={Wu, Bo and Yu, Shoubin and Chen, Zhenfang and Tenenbaum, Joshua B and Gan, Chuang},
  journal={arXiv preprint arXiv:2405.09711},
  year={2024}
}

@article{yi2019clevrer,
  title={Clevrer: Collision events for video representation and reasoning},
  author={Yi, Kexin and Gan, Chuang and Li, Yunzhu and Kohli, Pushmeet and Wu, Jiajun and Torralba, Antonio and Tenenbaum, Joshua B},
  journal={arXiv preprint arXiv:1910.01442},
  year={2019}
}

@inproceedings{xiao2021next,
  title={Next-qa: Next phase of question-answering to explaining temporal actions},
  author={Xiao, Junbin and Shang, Xindi and Yao, Angela and Chua, Tat-Seng},
  booktitle={Proceedings of the IEEE/CVF conference on computer vision and pattern recognition},
  pages={9777--9786},
  year={2021}
}

@inproceedings{wu2026spotlight,
  title={Spotlight and Shadow: Attention-Guided Dual-Anchor Introspective Decoding for MLLM Hallucination Mitigation},
  author={Wu, Yebo and Jin, Han and Guo, Zhijiang and Li, Li},
  booktitle={Findings of the Association for Computational Linguistics: ACL 2026},
  pages={13219--13233},
  year={2026}
}

@article{zhang2026hermes,
  title={HERMES: KV Cache as Hierarchical Memory for Efficient Streaming Video Understanding},
  author={Zhang, Haowei and Yang, Shudong and Fu, Jinlan and Ng, See-Kiong and Qiu, Xipeng},
  journal={arXiv preprint arXiv:2601.14724},
  year={2026}
}

@article{bai2025qwen3,
  title={Qwen3-vl technical report},
  author={Bai, Shuai and Cai, Yuxuan and Chen, Ruizhe and Chen, Keqin and Chen, Xionghui and Cheng, Zesen and Deng, Lianghao and Ding, Wei and Gao, Chang and Ge, Chunjiang and others},
  journal={arXiv preprint arXiv:2511.21631},
  year={2025}
}

@article{tang2026onevision,
  title={OneVision-Encoder: Codec-Aligned Sparsity as a Foundational Principle for Multimodal Intelligence},
  author={Tang, Feilong and An, Xiang and Yan, Yunyao and Xie, Yin and Qin, Bin and Yang, Kaicheng and Shen, Yifei and Zhang, Yuanhan and Li, Chunyuan and Feng, Shikun and others},
  journal={arXiv preprint arXiv:2602.08683},
  year={2026}
}

@inproceedings{zhou2024empirical,
  title={An empirical study on parameter-efficient fine-tuning for multimodal large language models},
  author={Zhou, Xiongtao and He, Jie and Ke, Yuhua and Zhu, Guangyao and Guti{\'e}rrez-Basulto, V{\'\i}ctor and Pan, Jeff},
  booktitle={Findings of the Association for Computational Linguistics: ACL 2024},
  pages={10057--10084},
  year={2024}
}

@inproceedings{xie2025adadare,
  title={AdaDARE-gamma: Balancing Stability and Plasticity in Multi-modal LLMs through Efficient Adaptation},
  author={Xie, Jingyi and Yang, Jintao and Luo, Zhunchen and Cao, Yunbo and Gao, Qiang and Zhang, Mengyuan and Hu, Wenpeng},
  booktitle={Proceedings of the Computer Vision and Pattern Recognition Conference},
  pages={19758--19768},
  year={2025}
}

@inproceedings{yao2025timechat,
  title={Timechat-online: 80\% visual tokens are naturally redundant in streaming videos},
  author={Yao, Linli and Li, Yicheng and Wei, Yuancheng and Li, Lei and Ren, Shuhuai and Liu, Yuanxin and Ouyang, Kun and Wang, Lean and Li, Shicheng and Li, Sida and others},
  booktitle={Proceedings of the 33rd ACM International Conference on Multimedia},
  pages={10807--10816},
  year={2025}
}

@article{qian2024streaming,
  title={Streaming long video understanding with large language models},
  author={Qian, Rui and Dong, Xiaoyi and Zhang, Pan and Zang, Yuhang and Ding, Shuangrui and Lin, Dahua and Wang, Jiaqi},
  journal={Advances in Neural Information Processing Systems},
  volume={37},
  pages={119336--119360},
  year={2024}
}

@article{black2024pi_0,
  title={$\pi_0$: A Vision-Language-Action Flow Model for General Robot Control},
  author={Black, Kevin and Brown, Noah and Driess, Danny and Esmail, Adnan and Equi, Michael and Finn, Chelsea and Fusai, Niccolo and Groom, Lachy and Hausman, Karol and Ichter, Brian and others},
  journal={arXiv preprint arXiv:2410.24164},
  year={2024}
}

@inproceedings{fu2025orion,
  title={Orion: A holistic end-to-end autonomous driving framework by vision-language instructed action generation},
  author={Fu, Haoyu and Zhang, Diankun and Zhao, Zongchuang and Cui, Jianfeng and Liang, Dingkang and Zhang, Chong and Zhang, Dingyuan and Xie, Hongwei and Wang, Bing and Bai, Xiang},
  booktitle={Proceedings of the IEEE/CVF International Conference on Computer Vision},
  pages={24823--24834},
  year={2025}
}

@inproceedings{li2024mvbench,
  title={Mvbench: A comprehensive multi-modal video understanding benchmark},
  author={Li, Kunchang and Wang, Yali and He, Yinan and Li, Yizhuo and Wang, Yi and Liu, Yi and Wang, Zun and Xu, Jilan and Chen, Guo and Luo, Ping and others},
  booktitle={Proceedings of the IEEE/CVF Conference on Computer Vision and Pattern Recognition},
  pages={22195--22206},
  year={2024}
}

@inproceedings{zhou2025mlvu,
  title={Mlvu: Benchmarking multi-task long video understanding},
  author={Zhou, Junjie and Shu, Yan and Zhao, Bo and Wu, Boya and Liang, Zhengyang and Xiao, Shitao and Qin, Minghao and Yang, Xi and Xiong, Yongping and Zhang, Bo and others},
  booktitle={Proceedings of the IEEE/CVF Conference on Computer Vision and Pattern Recognition},
  pages={13691--13701},
  year={2025}
}

@inproceedings{fu2025video,
  title={Video-mme: The first-ever comprehensive evaluation benchmark of multi-modal llms in video analysis},
  author={Fu, Chaoyou and Dai, Yuhan and Luo, Yongdong and Li, Lei and Ren, Shuhuai and Zhang, Renrui and Wang, Zihan and Zhou, Chenyu and Shen, Yunhang and Zhang, Mengdan and others},
  booktitle={Proceedings of the IEEE/CVF conference on computer vision and pattern recognition},
  pages={24108--24118},
  year={2025}
}

@article{team2026qwen3,
  title={Qwen3.5-Omni Technical Report},
  author={Team, Qwen},
  journal={arXiv preprint arXiv:2604.15804},
  year={2026}
}

@article{xu2026more,
  title={More thinking, less seeing? assessing amplified hallucination in multimodal reasoning models},
  author={Liu, Chengzhi and Xu, Zhongxing and Wei, Qingyue and Wu, Juncheng and Zou, James and Wang, Xin Eric and Zhou, Yuyin and Liu, Sheng},
  journal={arXiv preprint arXiv:2505.21523},
  year={2025}
}

@article{tan2024cradle,
  title={Cradle: Empowering foundation agents towards general computer control},
  author={Tan, Weihao and Zhang, Wentao and Xu, Xinrun and Xia, Haochong and Ding, Ziluo and Li, Boyu and Zhou, Bohan and Yue, Junpeng and Jiang, Jiechuan and Li, Yewen and others},
  journal={arXiv preprint arXiv:2403.03186},
  year={2024}
}

@article{yu2026vismem,
  title={Vismem: Latent vision memory unlocks potential of vision-language models},
  author={Yu, Xinlei and Xu, Chengming and Zhang, Guibin and Chen, Zhangquan and Zhang, Yudong and He, Yongbo and Jiang, Peng-Tao and Zhang, Jiangning and Hu, Xiaobin and Yan, Shuicheng},
  journal={arXiv preprint arXiv:2511.11007},
  year={2025}
}

@article{huang2026persistent,
  title={Persistent Visual Memory: Sustaining Perception for Deep Generation in LVLMs},
  author={Huang, Siyuan and Qu, Xiaoye and Li, Yafu and Zhu, Tong and He, Zefeng and Fu, Muxin and Liu, Daizong and Zheng, Wei-Long and Cheng, Yu},
  journal={arXiv preprint arXiv:2605.00814},
  year={2026}
}

@article{luo2026narrow,
  title={From narrow to panoramic vision: Attention-guided cold-start reshapes multimodal reasoning},
  author={Luo, Ruilin and Shi, Chufan and Zhang, Yizhen and Yang, Cheng and Jiang, Songtao and Guan, Tongkun and Chen, Ruizhe and Chu, Ruihang and Wang, Peng and Yang, Mingkun and others},
  journal={arXiv preprint arXiv:2603.03825},
  year={2026}
}

\appendix

\section{Experimental Details}\label{app:experimental_details}

\subsection{Benchmark Descriptions}\label{app:benchmark_descriptions}

The entries below summarize the scale and task organization of each
evaluation benchmark.
\begin{itemize}
    \item \textbf{StreamingBench}~\citep{lin2026streamingbench} contains $900$
    videos and $4{,}500$ human-curated question--answer pairs. Its $18$ tasks are
    organized into real-time visual understanding, omni-source understanding,
    and contextual understanding, with five questions associated with
    distinct timestamps in each video. The real-time visual understanding
    subset contains ten tasks.

    \item \textbf{MVBench}~\citep{li2024mvbench} comprises $20$
    multiple-choice video tasks organized under a static-to-dynamic taxonomy,
    including action order, object interaction, motion, and scene transition.

    \item \textbf{MLVU}~\citep{zhou2025mlvu} comprises long videos from
    several genres, including movies, surveillance footage, egocentric
    recordings, and cartoons. It defines seven multiple-choice tasks, and
    M-Avg denotes the mean of their task-level accuracies.

    \item \textbf{VideoMME}~\citep{fu2025video} contains $900$ videos and
    $2{,}700$ multiple-choice questions, totaling approximately $254$ hours. The
    videos cover six visual domains and $30$ subfields, span $11$ seconds to one
    hour, and are partitioned into short, medium, and long subsets.
\end{itemize}

\subsection{Baseline Setup}\label{app:baseline_setup}

The entries below summarize the defining mechanism of each comparison
baseline.
\begin{itemize}
    \item \textbf{VideoLLM-Online}~\citep{chen2024videollm} uses the
    Learning-In-Video-Stream (LIVE) framework, which combines a streaming
    language-modeling objective, offline-to-streaming dialogue construction,
    and interleaved frame processing and response generation.

    \item \textbf{Dispider}~\citep{qian2025dispider} separates active
    streaming interaction into perception, decision, and reaction. A
    lightweight module monitors the stream and determines when to respond;
    response generation and subsequent video perception are executed
    asynchronously.

    \item \textbf{TimeChat-Online}~\citep{yao2025timechat} uses Differential
    Token Drop to retain visual tokens associated with temporal changes while
    removing redundant content across adjacent frames. It preserves the
    original spatiotemporal positions of retained tokens and supports online
    interaction through continuous frame processing.

    \item \textbf{StreamForest}~\citep{zeng2026streamforest} organizes
    long-term visual history into event-level trees through a Persistent Event
    Memory Forest. Temporal distance, content similarity, and merge frequency
    determine event consolidation, while a spatiotemporal window
    stores recent observations.

    \item \textbf{ReKV}~\citep{di2025streaming} stores historical video KV
    caches in an external memory and retrieves query-relevant entries when a
    question arrives. This decouples streaming video encoding from question
    answering while keeping the active language-model context compact.

    \item \textbf{LiveVLM}~\citep{ning2025livevlm} is a training-free,
    query-agnostic method that compresses streaming video KV caches with
    Vision Sink Bucketing, using token importance and temporal coverage as
    retention criteria. Position-agnostic KV Retrieval combines retrieved
    long-term cache pages with a sliding window of recent KVs.

    \item \textbf{StreamMem}~\citep{yang2025streammem} is a training-free,
    query-agnostic method that incrementally encodes video while maintaining
    a fixed-size KV cache. It uses attention from generic chat-template tokens
    to prune visual KVs and frame-wise merging to retain compact prototype
    representations of observed frames.

    \item \textbf{InfiniPot-V}~\citep{kim2026infinipot} is training-free and
    query-agnostic, maintaining a length-independent KV-cache budget during
    video encoding. Temporal-axis Redundancy removes repetitive cache
    entries, and Value-Norm ranking determines the retained entries.

    \item \textbf{StreamingTOM}~\citep{chen2026streamingtom} is a
    training-free method that manages visual tokens before and after the
    language model. Causal Temporal Reduction selects a fixed-budget subset
    from each frame, while Online Quantized Memory stores historical groups
    in $4$-bit form and retrieves selected groups on demand.

    \item \textbf{HERMES}~\citep{zhang2026hermes} is a training-free method
    that treats the Video-LLM KV cache as hierarchical memory containing
    visual information at multiple granularities. The cache is updated
    incrementally and reused when a question arrives without a separate
    auxiliary query-time processor.

    \item \textbf{FluxMem}~\citep{xie2026fluxmem} is a training-free method
    that organizes visual context into short-, mid-, and long-term memory.
    Temporal Adjacency Selection removes redundant tokens across adjacent
    frames, while Spatial Domain Consolidation merges repetitive spatial
    regions using adaptive, per-frame thresholds.

    \item \textbf{SimpleStream}~\citep{shen2026simple} retains a fixed-size
    sliding window of recent frames and supplies the window directly to an
    off-the-shelf VLM. It uses neither explicit long-term memory nor
    retrieval, compression, or streaming-specific training.

    \item \textbf{CausalMem}~\citep{song2026towards} is a training-free
    method that maintains a fixed-budget visual memory bank. It estimates
    token redundancy with an incrementally updated semantic basis and balances
    semantic novelty with temporal recency when selecting retained tokens.
\end{itemize}

\subsection{Training Data}\label{app:training_data}

In this section, we provide detailed descriptions of each dataset used in our experiment.

\paragraph{NeXT-QA.}
NeXT-QA~\citep{xiao2021next} is a video question-answering dataset designed to move beyond scene description toward causal and temporal action reasoning. It contains $5{,}440$ videos with approximately $52$K manually annotated question--answer pairs, organized into causal, temporal, and descriptive categories. The benchmark supports both multiple-choice QA, with five answer candidates per question, and open-ended QA requiring a short generated answer. We use its official training partition, whose daily-activity videos and manually written questions provide supervision for event causes, temporal order, and object interactions.

\paragraph{COIN.}
COIN~\citep{tang2019coin} is a large-scale instructional-video dataset containing $11{,}827$ videos across $180$ tasks and $12$ domains of daily life. Its annotation hierarchy organizes videos by domain and task, and each video is labeled with a sequence of step descriptions together with their temporal boundaries. We use the official training partition and convert its temporally localized procedure annotations into question-conditioned training examples. This source provides supervision in which the ordering and dependency of procedural steps are central to understanding the video.

\paragraph{STAR.}
STAR~\citep{wu2024star} is a diagnostic benchmark for situated reasoning in real-world videos, comprising approximately $22$K situation clips and $60$K questions. It defines four question types---interaction, sequence, prediction, and feasibility---over videos involving human actions and human--object interactions. Each situation is represented by a hypergraph connecting entities and relations such as people, objects, actions, and their relationships; questions and answers are procedurally generated, and the reasoning process for each question is specified by a functional program grounded in that hypergraph. We use the official training partition to provide supervision for action interaction, temporal progression, future prediction, and feasible-action reasoning.

\paragraph{CLEVRER.}
CLEVRER~\citep{yi2019clevrer} is a controlled diagnostic dataset for temporal and causal reasoning about physical events. It contains $20{,}000$ synthetic videos of moving and colliding objects and more than $300$K questions spanning descriptive, explanatory, predictive, and counterfactual reasoning. The dataset provides ground-truth object motion traces and event histories, and each question is paired with a functional program that expresses its underlying logic. We use the official training partition, which supplies explicit supervision for object dynamics, collision causality, future outcomes, and counterfactual physical events.

\paragraph{LLaVA-Video-178K.}
LLaVA-Video-178K~\citep{zhang2024llava} is a synthetic video instruction-following dataset built from $178{,}510$ videos ranging from zero to three minutes. The full release contains approximately $1.3$M instruction-following samples, including $178$K detailed captions, $960$K open-ended QA pairs, and $196$K multiple-choice QA pairs. Its annotations were constructed with GPT-4o and human effort, with question types derived from a survey of public video captioning and QA benchmarks. The dataset complements the more structured reasoning sources above with diverse real-world videos and broader instruction formats.

\subsection{Implementation Details}\label{app:implementation_details}

The hyperparameter configuration used for all benchmark comparisons, sensitivity studies, ablations, mechanistic VAS analyses, and counterfactual grounding experiments is summarized in \Cref{tab:implementation_hyperparameters}.

\begin{table}[t]
    \centering
    \small
    \setlength{\tabcolsep}{4.5pt}
    \renewcommand{\arraystretch}{1.05}
    \caption{Hyperparameter settings used in our experiments.}
    \label{tab:implementation_hyperparameters}
    \begin{tabular}{@{}p{0.56\columnwidth}p{0.34\columnwidth}@{}}
        \toprule
        \textbf{Parameter} & \textbf{Setting} \\
        \midrule
        Answering backbone & Qwen3.5-9B \\
        Memory compressor & Qwen3.5-9B with LoRA \\
        Answering-backbone update & Frozen \\
        Compressor adaptation & LoRA, rank 16 \\
        Trainable modules & Compressor LoRA, latent queries, and projections \\
        Training objective & Token-level cross-entropy \\
        Supervision scheme & Teacher forcing \\
        Training hardware & 4$\times$ NVIDIA L20X GPUs \\
        Evaluation frame rate & 1 FPS \\
        GOP length $T$ & 4 frames \\
        Mid-term memory capacity & 32 frames \\
        P-frame retention ratio $\rho$ & 0.5 \\
        Latent tokens per injection $L$ & 32 \\
        Long-term memory capacity $C$ & 96 GOPs \\
        Merge operator $\mathcal{F}_{\mathrm{merge}}$ & Similarity-guided anchor merge \\
        Retrieved GOPs per injection $K$ & 4 \\
        VAS threshold $\tau$ & 0.10 \\
        Maximum dynamic injections & 5 \\
        Minimum injection interval & 8 decoding steps \\
        \bottomrule
    \end{tabular}
\end{table}

\paragraph{Teacher Label Construction.}
We use Qwen3.5-27B as a video-conditioned teacher to construct evidence-augmented SFT labels. Given a processed video and its question context, the teacher produces two fields: \texttt{Evidence}, a concise factual paragraph grounded in visible content, and \texttt{Answer}, the predicted answer. The prompt excludes reasoning traces, answer comparisons, uncertainty, conclusions, and meta commentary from the evidence.

As shown in \Cref{alg:teacher_label_construction}, generation proceeds from weak to stronger supervision. We first perform answer-unconditioned greedy decoding, followed, when necessary, by answer-unconditioned nucleus sampling. Only unresolved examples enter the final stage, which supplies the reference answer to guide evidence extraction while prohibiting disclosure of this supervision signal. A candidate is valid only if it satisfies the output and evidence constraints and predicts the reference answer. Within each sampling stage, we select the valid candidate with the highest mean token log-probability. Examples without a valid candidate after all stages are recorded as failures.

\begin{algorithm}[t]
    \caption{Staged construction of video-evidence SFT labels.}
    \label{alg:teacher_label_construction}
    \begin{algorithmic}[1]
        \REQUIRE Processed video $V$, question context $q$, reference answer $y$, teacher $\mathcal{T}$, sampling configurations $\theta_{\mathrm{u}}$ and $\theta_{\mathrm{c}}$
        \ENSURE Evidence-augmented label $\ell$ or failure
        \STATE $\mathcal{C}_{\mathrm{g}}\leftarrow\textsc{Greedy}(\mathcal{T},V,q)$
        \STATE $c^{\star}\leftarrow\textsc{BestValid}(\mathcal{C}_{\mathrm{g}},y)$
        \IF{$c^{\star}=\varnothing$}
            \STATE $\mathcal{C}_{\mathrm{u}}\leftarrow\textsc{Nucleus}(\mathcal{T},V,q;\theta_{\mathrm{u}})$
            \STATE $c^{\star}\leftarrow\textsc{BestValid}(\mathcal{C}_{\mathrm{u}},y)$
        \ENDIF
        \IF{$c^{\star}=\varnothing$}
            \STATE $\mathcal{C}_{\mathrm{c}}\leftarrow\textsc{Nucleus}(\mathcal{T},V,q,y;\theta_{\mathrm{c}})$
            \STATE $c^{\star}\leftarrow\textsc{BestValid}(\mathcal{C}_{\mathrm{c}},y)$
        \ENDIF
        \IF{$c^{\star}=\varnothing$}
            \STATE \textbf{fail}
        \ENDIF
        \STATE \RETURN $\ell\leftarrow\textsc{Label}(c^{\star})$
    \end{algorithmic}
\end{algorithm}

\section{Technical Details}
\label{app:preliminary}
\label{app:analysis_details}

\paragraph{Notation.}
The notation below is grouped by pipeline stage.
Bold uppercase symbols denote frame- or sequence-level tensors, bold lowercase
symbols denote vectors, calligraphic symbols denote memories, sets, and
operators, and italic symbols denote scalar indices and budgets.
\begin{itemize}
    \item \textbf{Video stream and mid-term memory.}
    $g$ indexes groups of pictures (GOPs), and $T$ is the number of frames
    per GOP. The $g$-th GOP is
    $\mathcal{G}^{g}=\{\mathbf{X}^{g}_{0},\ldots,
    \mathbf{X}^{g}_{T-1}\}$, where $f$ indexes frames,
    $\mathbf{X}^{g}_{0}$ is the I-frame, and $\mathbf{X}^{g}_{f}$ for $f>0$
    is a P-frame. $\mathbf{u}$ denotes a spatial location;
    $\mathbf{R}^{g}_{f}(\mathbf{u})$ is the RGB residual at that location.
    For patch $i$, $\mathcal{P}_{i}$ is its pixel set and
    $s^{g}_{f,i}$ is its residual score. $N$ is the number of I-frame
    patches, $\rho$ is the P-frame retention ratio, and
    $\mathcal{S}^{g}_{f}$ is the set of $\lceil\rho N\rceil$ retained
    P-frame patches. $\widehat{\mathcal{G}}^{g}$ denotes the sparse GOP
    representation stored in the mid-term memory
    $\mathcal{M}_{\mathrm{M}}$.

    \item \textbf{Encoded GOPs and long-term memory.}
    $\mathcal{E}_{v}$ denotes the visual encoder. The encoded representation
    of GOP $g$ is
    \[
    \mathbf{H}^{g}=[\mathbf{h}^{g}_{1},\ldots,
    \mathbf{h}^{g}_{N_g}]^{\top}\in\mathbb{R}^{N_g\times d_v},
    \]
    where $N_g$ is its number of
    retained visual tokens and $d_v$ is the visual hidden dimension.
    The long-term memory is
    $\mathcal{M}_{\mathrm{L}}=[\mathbf{H}^{(1)},\ldots,
    \mathbf{H}^{(n)}]$, where $C$ is its capacity in GOPs and
    $0\le n\le C$. $\mathbf{H}^{(j)}=
    [\mathbf{I}_{j},\mathbf{P}^{1}_{j},\ldots,
    \mathbf{P}^{T-1}_{j}]$ is the $j$-th stored GOP, with
    $\mathbf{I}_{j}$ and $\mathbf{P}^{f}_{j}$ denoting its I-frame and
    P-frame token sequences; $\mathbf{H}^{\mathrm{new}}$ denotes an
    incoming GOP. $s_j$ is the adjacent-GOP I-frame similarity,
    $p$ is the index of the most similar adjacent pair,
    $\mathcal{F}_{\mathrm{merge}}$ is the GOP merge operator, and
    $\overline{\mathbf{H}}^{(p)}$ is the merged representation.

    \item \textbf{Attention and retrieval.}
    $t$ indexes autoregressive generation steps, $c$ indexes context
    positions, and $\ell$ and $h$ index layers and attention heads.
    $A_{t,c}^{\ell,h}$ is the corresponding normalized attention weight;
    $N_{\ell}$ and $N_h$ are the numbers of layers and heads, and
    $\mathcal{V}_{t}$ is the set of accessible visual-token positions.
    $\operatorname{VAS}_{t}$ is their layer- and head-averaged attention
    mass, and $\tau$ is the retrieval threshold. $\mathcal{P}_{v}$ denotes
    the visual projector, $\mathcal{Q}_{t}$ the query-token set, and
    $\mathbf{e}_{u}$ the embedding of query token $u$;
    $\mathbf{q}_{t}$ is the pooled query. For frame $f$ in stored GOP $j$,
    $\mathbf{h}^{(j)}_{f,i}$ is its $i$-th valid visual token,
    $N_{j,f}$ is the number of valid tokens, and $\mathbf{k}_{j,f}$ is the
    pooled frame key. $r_{t,j}$ is the relevance score of GOP $j$ and
    $\mathcal{I}_{t}$ is the chronologically sorted index set of at most
    $K$ retrieved GOPs; $\mathbf{H}^{(\mathcal{I}_{t})}$ denotes their
    concatenation. $\mathcal{C}_{\phi}$ is the compressor parameterized by
    $\phi$, $L$ is the number of learned queries, $\mathbf{L}$ denotes
    those queries, and $\mathbf{Z}_{t}$ is the resulting length-$L$ visual
    latent sequence.
\end{itemize}

\paragraph{Disambiguation.}
Superscript $g$ indexes GOPs at arrival, whereas parenthesized superscript
$(j)$ indexes their order in long-term memory. The symbols
$s^{g}_{f,i}$ and $s_j$ denote patch residual and adjacent-GOP similarity,
respectively. The counts $N$, $N_g$, and $N_{j,f}$ refer to the full
I-frame patch/token count, the retained-token count of an encoded GOP, and
the valid-token count of a stored frame. The symbols $\mathcal{P}_{i}$,
$\mathbf{P}^{f}_{j}$, and $\mathcal{P}_{v}$ denote a patch pixel set, a
P-frame token sequence, and the visual projector. $\mathcal{I}$ in the patch
selection constraint is a dummy candidate set, whereas $\mathcal{I}_{t}$ is
the retrieved GOP index set. $L$ is a count and $\mathbf{L}$ is the
learned-query sequence. The symbol $\rho$ denotes the P-frame retention ratio;
Spearman's $\rho$ in the analysis denotes the rank-correlation
coefficient.

\paragraph{Causal Setting.}
At a question timestamp, the input is restricted to the question and the
video frames observed up to that timestamp. Future frames are excluded. The
memory states $\mathcal{M}_{\mathrm{M}}$ and $\mathcal{M}_{\mathrm{L}}$
contain only representations of the observed video prefix.

\subsection{Visual Attention Score Computation}\label{app:vas_computation}

The Visual Attention Score (VAS) measures how much attention the token currently being decoded assigns to accessible visual evidence. At generation step $t$, let $A_{t,c}^{\ell,h}$ denote the normalized causal attention weight from the current query position to context position $c$ at layer $\ell$ and head $h$. Let $\mathcal{V}_{t}$ contain the indices of all visual context tokens available at that step, including the visual prefix and any long-term memory latents inserted earlier in generation. We first sum the attention weights over $\mathcal{V}_{t}$ for each layer--head pair and then average these visual attention masses across all $N_{\ell}$ layers and $N_{h}$ heads:
\begin{equation}
\operatorname{VAS}_{t}
=
\frac{1}{N_{\ell}N_{h}}
\sum_{\ell=1}^{N_{\ell}}
\sum_{h=1}^{N_{h}}
\sum_{c\in\mathcal{V}_{t}}
A_{t,c}^{\ell,h}.
\label{eq:supp_vas}
\end{equation}

\begin{figure*}[t]
    \centering
    \includegraphics[width=0.86\textwidth,trim=0 2bp 0 0,clip]{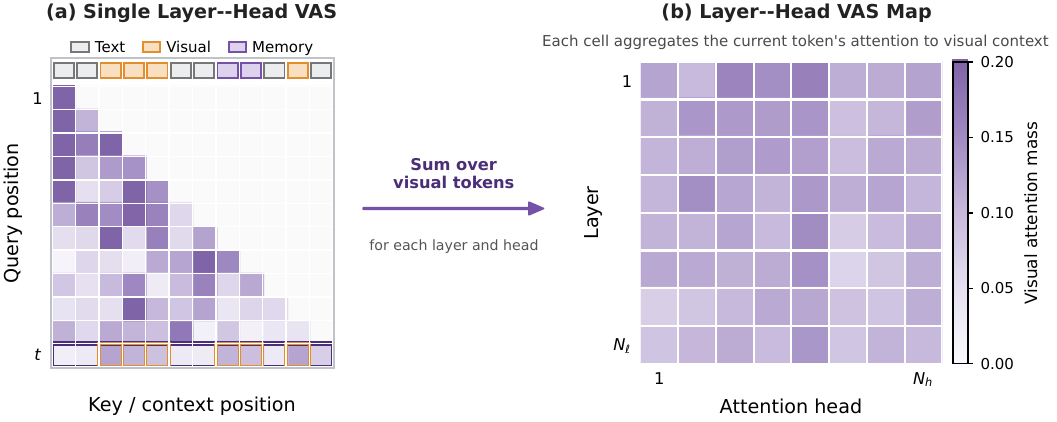}
    \caption{Computation of the Visual Attention Score at generation step $t$. Left: for one layer and attention head, VAS aggregates the current token's normalized attention weights over the visual-prefix and inserted-memory positions. Right: repeating this aggregation for every layer--head pair produces a visual-attention-mass map, whose mean is $\operatorname{VAS}_{t}$.}
    \label{fig:supp_vas_computation}
\end{figure*}

As illustrated in \Cref{fig:supp_vas_computation}, the inner sum in \Cref{eq:supp_vas} reduces one attention row to the mass assigned specifically to visual context, rather than to previously generated text. Averaging the resulting layer--head map yields a single token-level score in $[0,1]$. A high VAS indicates that the current decoding state is actively using available visual evidence, whereas a low VAS indicates that linguistic context dominates the attention distribution. \ourmethod uses this score online and triggers retrieval when $\operatorname{VAS}_{t}<\tau$, so the decision to expose additional long-term memory is tied directly to the model's current visual reliance.

\subsection{Counterfactual Intervention}\label{app:counterfactual_intervention}

A counterfactual asks how an outcome would change under a specific alternative condition while holding all other relevant factors fixed. Unlike an ordinary comparison between two independently configured systems, a counterfactual intervention changes only the variable whose causal contribution is being tested. The observed condition serves as the factual reference, and the intervened condition represents what would have happened to the same example had that variable taken a controlled alternative value. Differences between the paired outcomes can therefore be attributed to the intervention under the stated controls.

In our grounding analysis, the factual condition supplies each sample with its own matched long-term visual memory. The counterfactual condition replaces only that memory with the memory of another sample, while keeping the query, raw visual tokens, model parameters, answer options, decoding procedure, and visual-token budget unchanged. This intervention preserves the presence and amount of memory but breaks its semantic correspondence with the current video. Comparing matched and shuffled memory therefore tests whether VAS tracks the use of relevant visual evidence, rather than merely responding to the existence of additional visual tokens. We measure the effect through paired changes in VAS and in the correct-option log-odds.

\section{Extra Results}\label{app:extra_results}
\subsection{Additional Benchmark Results}\label{app:additional_benchmark_results}
Beyond the main-paper comparison, we further adapt a selected subset of baselines to use Qwen3.5-9B as their shared backbone. We evaluate these adapted methods on StreamingBench, MVBench, and VideoMME. This controlled comparison complements the main results by reducing differences in backbone capability and more directly comparing the streaming-memory strategies of the evaluated methods. The results are reported in \Cref{tab:supp_qwen35_9b_results}.

\par\smallskip
\noindent\begin{minipage}{\textwidth}
    \centering
    \small
    \setlength{\tabcolsep}{3.5pt}
    \renewcommand{\arraystretch}{1.08}
    \definecolor{tblgroup}{gray}{0.95}
    \definecolor{tblours}{gray}{0.88}
    \captionof{table}{
    Additional comparison of selected baselines adapted to the Qwen3.5-9B backbone on StreamingBench, MVBench, and VideoMME. All entries are accuracy (\%). The best and second-best results are in \textbf{bold} and \underline{underlined}, respectively.
    }
    \label{tab:supp_qwen35_9b_results}
    \resizebox{0.92\textwidth}{!}{%
    \begin{tabular}{@{}l cccc cc@{}}
    \toprule
    \multirow{2}{*}{\textbf{Method}} &
    \multicolumn{4}{c}{\textbf{StreamingBench}} &
    \multirow{2}{*}{\textbf{MVBench}} &
    \multirow{2}{*}{\textbf{VideoMME}} \\
    \cmidrule(lr){2-5}
    & \textbf{CU} & \textbf{OSU} & \textbf{RTVU} & \textbf{Overall}
    & & \\
    \midrule
    \rowcolor{tblgroup}
    Qwen3.5-9B & 37.60 & 43.40 & 76.34 & 63.26 & 67.43 & 70.74 \\
    ReKV~\cite{di2025streaming} & 35.20 & 46.00 & 78.26 & 65.78 & 68.75 & \underline{71.89} \\
    LiveVLM~\cite{ning2025livevlm} & 38.00 & 45.00 & 79.14 & 65.46 & 68.77 & 69.26 \\
    InfiniPot-V~\cite{kim2026infinipot} & 35.10 & 45.90 & 79.82 & 65.78 & 67.45 & 70.89 \\
    StreamingTOM~\cite{chen2026streamingtom} & 37.60 & 46.30 & \underline{80.46} & \underline{66.56} & \textbf{69.00} & 66.30 \\
    SimpleStream~\cite{shen2026simple} & \underline{38.20} & \textbf{47.70} & 79.02 & 66.08 & 67.43 & 63.15 \\
    \rowcolor{tblours}
    \textbf{\ourmethod{}-9B (ours)} & \textbf{39.20} & \underline{47.50} & \textbf{81.55} & \textbf{67.73} & \underline{68.90} & \textbf{73.52} \\
    \bottomrule
    \end{tabular}%
    }
\end{minipage}
\par\medskip

\textbf{\ourmethod Leads the Controlled Streaming Comparison.}
As shown in \Cref{tab:supp_qwen35_9b_results}, \ourmethod achieves the best StreamingBench overall accuracy of $67.73\%$, outperforming the strongest competing memory strategy, StreamingTOM, by $1.17$ percentage points. The advantage is particularly clear on RTVU, where \ourmethod reaches $81.55\%$ and exceeds StreamingTOM by $1.09$ points. It also obtains the highest CU score of $39.20\%$, while its OSU score of $47.50\%$ is within $0.20$ points of the best result. Relative to the vanilla Qwen3.5-9B backbone, \ourmethod improves CU, OSU, RTVU, and overall accuracy by $1.60$, $4.10$, $5.21$, and $4.47$ points, respectively. These gains under a shared answering backbone indicate that the improvements arise from the proposed streaming-memory design rather than from differences in backbone capacity.

\textbf{\ourmethod Generalizes Strongly to Offline Video Understanding.}
The same controlled comparison further demonstrates that \ourmethod generalizes strongly beyond StreamingBench. It achieves $68.90\%$ on MVBench and the highest VideoMME accuracy of $73.52\%$, surpassing the second-best ReKV result by $1.63$ points. Overall, \ourmethod ranks first on four of the six reported metrics and second on the remaining two. This consistency demonstrates that its two-stage memory preserves both recent details and long-range evidence while maintaining the general video-understanding capability of the frozen backbone.

\Needspace{8\baselineskip}
\subsection{Single-Sample Attention Dynamics}\label{app:single_sample_vas}

\Cref{fig:supp_single_sample_vas} examines the prospective-reasoning question ``What might the speaker discuss next?'' The video ends with a rotating disc divided into colored regions, and the correct continuation concerns the probability of the disc landing on the blue region. Visual-prefix baselines exhibit pronounced attention drift: after an early rise, their VAS falls steadily as generation proceeds. In contrast, \ourmethod maintains a mean VAS of $0.152$, compared with $0.081$ for the strongest baseline on this sample, and remains above every baseline for $89.2\%$ of normalized generation.

\par\smallskip
\noindent\begin{minipage}{\textwidth}
    \centering
    \includegraphics[width=0.72\textwidth,trim=0 25bp 0 0,clip]{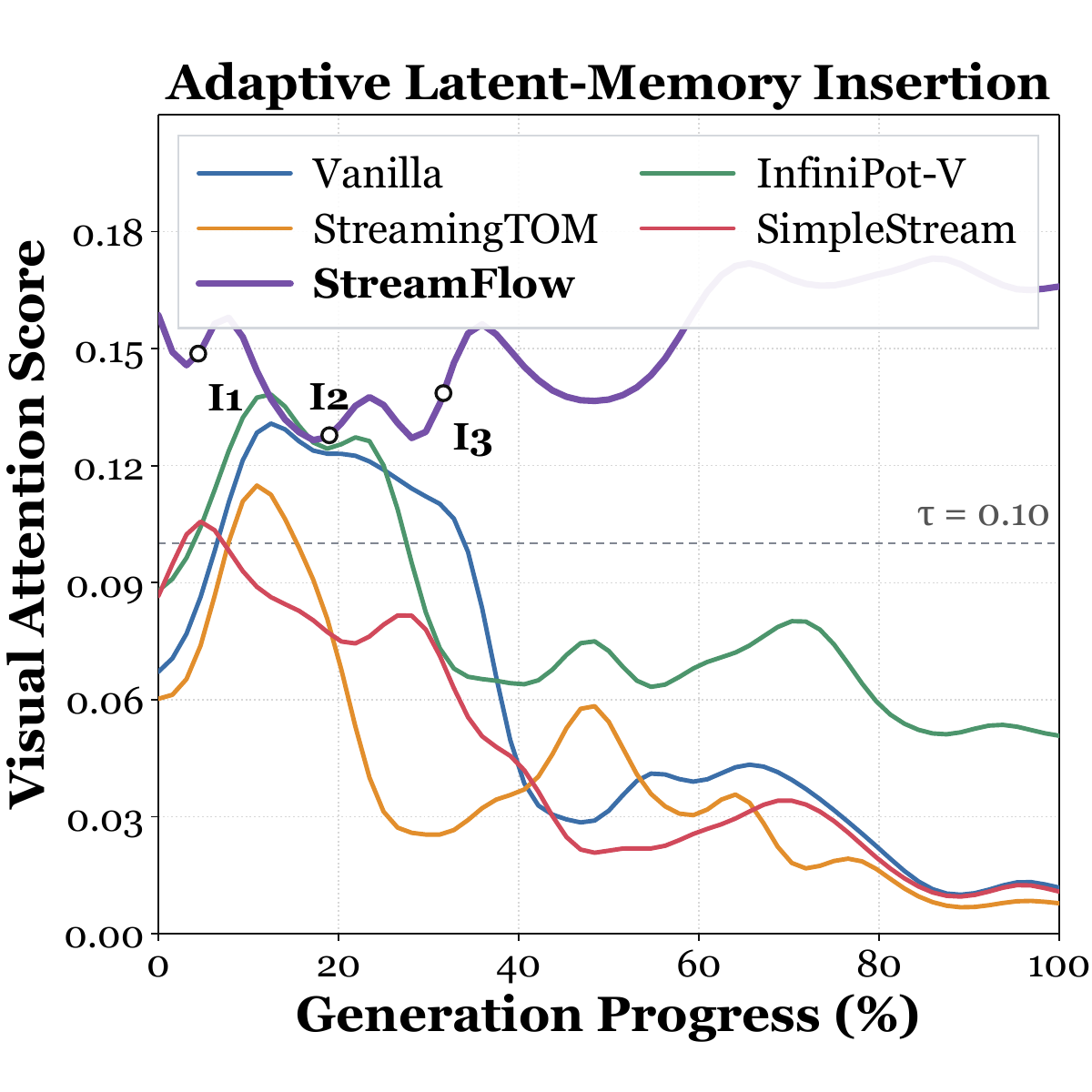}
    \captionof{figure}{Generation-time VAS trajectories for an illustrative RTVU prospective-reasoning sample. Curves are resampled by normalized generation progress and smoothed for visualization. $I_1$--$I_3$ mark the three dynamic long-term-memory insertions made by \ourmethod; the dashed line denotes the trigger threshold $\tau=0.10$.}
    \label{fig:supp_single_sample_vas}
\end{minipage}
\par\medskip

The three markers show how the online policy responds to this drift. VAS reaches $0.100$, $0.086$, and $0.086$ immediately before $I_1$, $I_2$, and $I_3$, respectively. After the corresponding memories are inserted, VAS on the next generated text token rises to $0.203$, $0.152$, and $0.177$, giving absolute lifts of $0.103$, $0.066$, and $0.091$. Thus, each retrieval event is followed by a local recovery of attention to accessible visual evidence, while repeated insertions keep the trajectory above the visual-prefix baselines later in decoding. This example illustrates the token-level behavior underlying the aggregate trend in the main paper.

\Needspace{20\baselineskip}
\subsection{Sensitivity Analysis}\label{app:sensitivity_analysis}

\subsubsection{Latent Length}\label{app:latent_length_sensitivity}
We evaluate the number of latent tokens produced per memory insertion on MLVU while holding the remaining configuration fixed. As shown in \Cref{fig:supp_latent_length_sensitivity}, accuracy increases monotonically from $71.16\%$ at $L=4$ to $75.83\%$ at $L=64$. The largest gain occurs between $L=8$ and $L=16$ ($+2.25$ points), after which the increments narrow to $+0.89$ points from $L=16$ to $L=32$ and $+0.49$ points from $L=32$ to $L=64$. Consequently, the default $L=32$ improves over $L=4$ by $4.18$ points and captures $89.5\%$ of the total improvement observed over the tested range while using half as many latent tokens as $L=64$. This saturation establishes $L=32$ as a practical operating point that combines $75.34\%$ accuracy with a compact latent representation.

\par\smallskip
\noindent\begin{minipage}{\textwidth}
    \centering
    \includegraphics[width=0.72\textwidth,trim=0 7bp 0 0,clip]{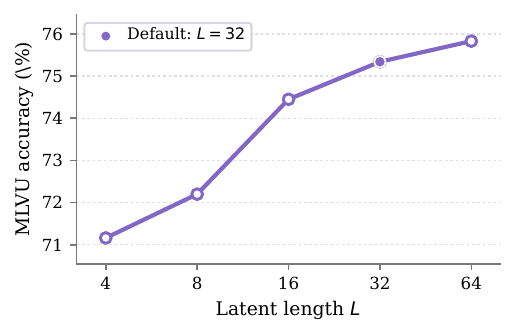}
    \captionof{figure}{Sensitivity to the latent length per memory insertion on MLVU. Accuracy improves monotonically as $L$ increases, with diminishing gains beyond the default $L=32$.}
    \label{fig:supp_latent_length_sensitivity}
\end{minipage}
\par\medskip

\subsubsection{P-Frame Retention Ratio}\label{app:retention_ratio_masks}

\paragraph{Selection Baselines.}
All policies use the same GOP-wide patch budget. Residual Top-$K$ (\ourmethod) retains the P-frame patches with the largest residuals from the I-frame. Random samples patches uniformly, Uniform Spatial selects approximately evenly spaced locations within each P-frame, and Uniform Frames retains complete P-frames at regular intervals while assigning any remaining budget to one additional frame.

\paragraph{Retention-Ratio Visualization.}
We visualize how the P-frame retention ratio changes the spatial support selected by dynamics-aware residual scoring. Each frame is partitioned into a $16\times28$ grid of $448$ patches. The I-frame is always retained in full, whereas each P-frame retains exactly $56$, $112$, $224$, or $336$ patches at $\rho=0.125$, $0.25$, $0.50$, or $0.75$, respectively. For each sample, all four masks use the same residual ranking, so the selected sets are strictly nested as the budget increases.

\Cref{fig:supp_retention_ratio_sample_128} shows that the portrait sequence concentrates its lowest-budget mask around the moving brush and the changing facial region. In \Cref{fig:supp_retention_ratio_sample_74}, the mask follows the presenter's changing pose while initially suppressing much of the static instruction board. \Cref{fig:supp_retention_ratio_sample_215} further shows that newly appearing words and graphics are retained together with the moving speaker. Across all three examples, the $12.5\%$ masks suppress broad temporally stable regions but preserve localized changes that distinguish each P-frame from its I-frame. The $25\%$ and default $50\%$ ratios expand coverage around the same dynamic evidence, rather than replacing it with unrelated patches, while the $75\%$ ratio mainly restores additional static context. Together, \Cref{fig:supp_retention_ratio_sample_128,fig:supp_retention_ratio_sample_74,fig:supp_retention_ratio_sample_215} show that increasing retention broadens contextual coverage, but the most salient temporal changes are already prioritized at lower budgets.

\Cref{fig:supp_retention_ratio_statistics} confirms that the qualitative behavior generalizes across retention budgets. At $12.5\%$ retention, \ourmethod captures $18.57\%$ of local motion and $26.19\%$ of appearance change, compared with at most $12.75\%$ and $12.74\%$, respectively, for the fixed-budget alternatives; it simultaneously rejects $99.44\%$ of independently identified static patches, versus at most $87.59\%$. At the default $50\%$ ratio, motion coverage and appearance-change coverage rise to $61.08\%$ and $71.96\%$, while static rejection remains $95.28\%$. Even at $75\%$ retention, \ourmethod retains $82.96\%$ of local motion and $90.05\%$ of appearance change while rejecting $87.69\%$ of static patches. The consistent separation across budgets shows that residual ranking allocates additional capacity to evolving regions before admitting redundant background, rather than obtaining its advantage only at the default operating point.

\par\smallskip
\noindent\begin{minipage}{\textwidth}
    \centering
    \includegraphics[width=0.76\textwidth,trim=0 2bp 0 0,clip]{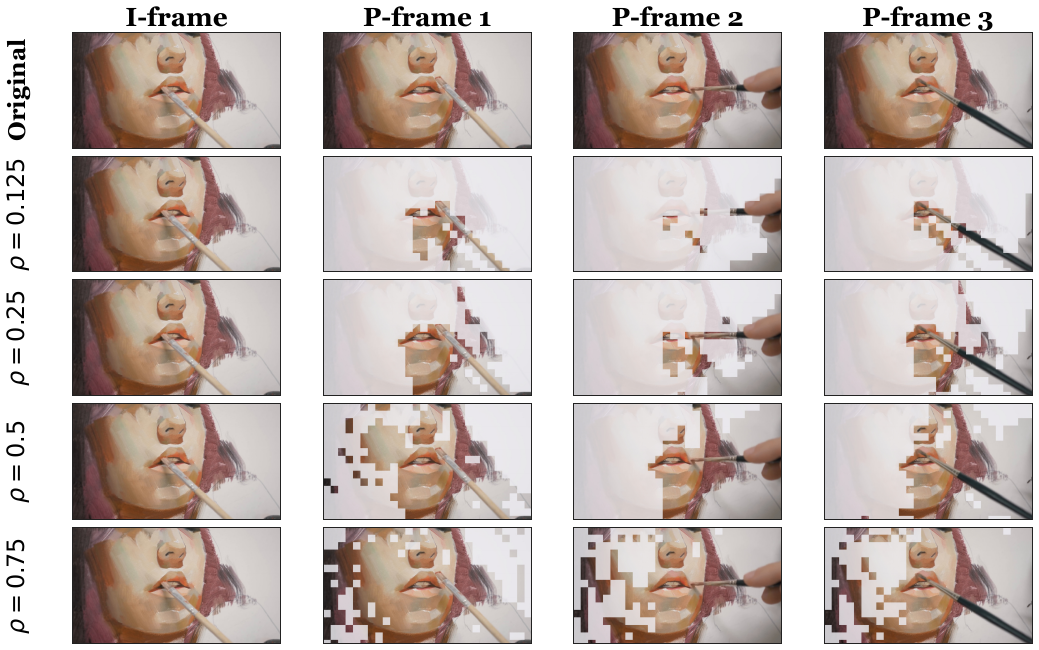}
    \captionof{figure}{Retention-ratio masks for a Clips Summarize sample involving a portrait being painted. At $\rho=0.125$, retained P-frame patches concentrate around the moving brush and the changing mouth and face regions. Larger ratios progressively restore the surrounding portrait and background while preserving earlier selections.}
    \label{fig:supp_retention_ratio_sample_128}
    \vspace{0.4em}
    \includegraphics[width=0.76\textwidth,trim=0 2bp 0 0,clip]{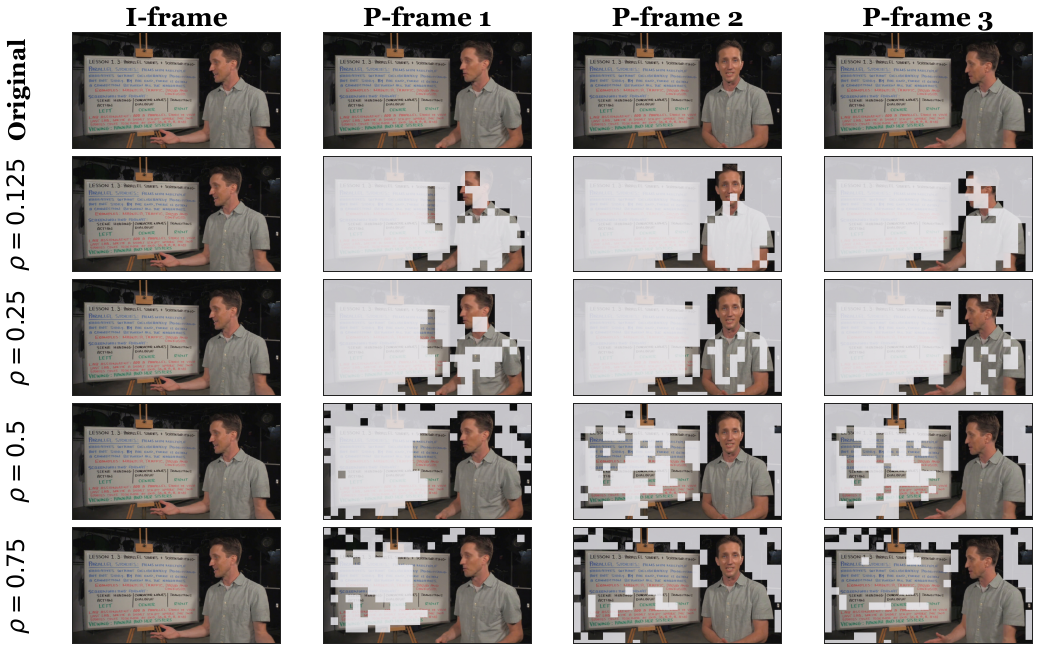}
    \captionof{figure}{Retention-ratio masks for a Counting sample with a moving presenter beside a largely static instruction board. The lowest budget emphasizes the presenter's changing pose and boundary regions; increasing $\rho$ adds the board and broader scene context.}
    \label{fig:supp_retention_ratio_sample_74}
\end{minipage}
\par\medskip

\par\smallskip
\noindent\begin{minipage}{\textwidth}
    \centering
    \includegraphics[width=0.76\textwidth,trim=0 2bp 0 0,clip]{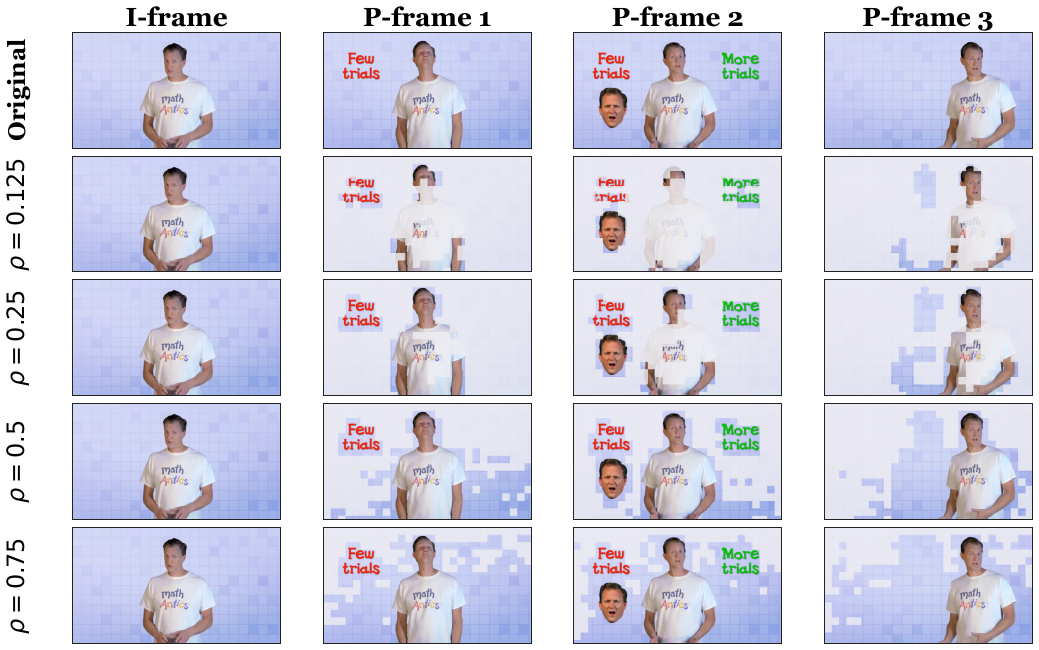}
    \captionof{figure}{Retention-ratio masks for a Prospective Reasoning sample. At $\rho=0.125$, residual selection preserves the moving speaker together with newly appearing textual and graphic elements. Higher-ratio masks expand around these regions and recover progressively more of the background.}
    \label{fig:supp_retention_ratio_sample_215}
    \vspace{0.6em}
    \includegraphics[width=0.88\textwidth,trim=0 3bp 0 0,clip]{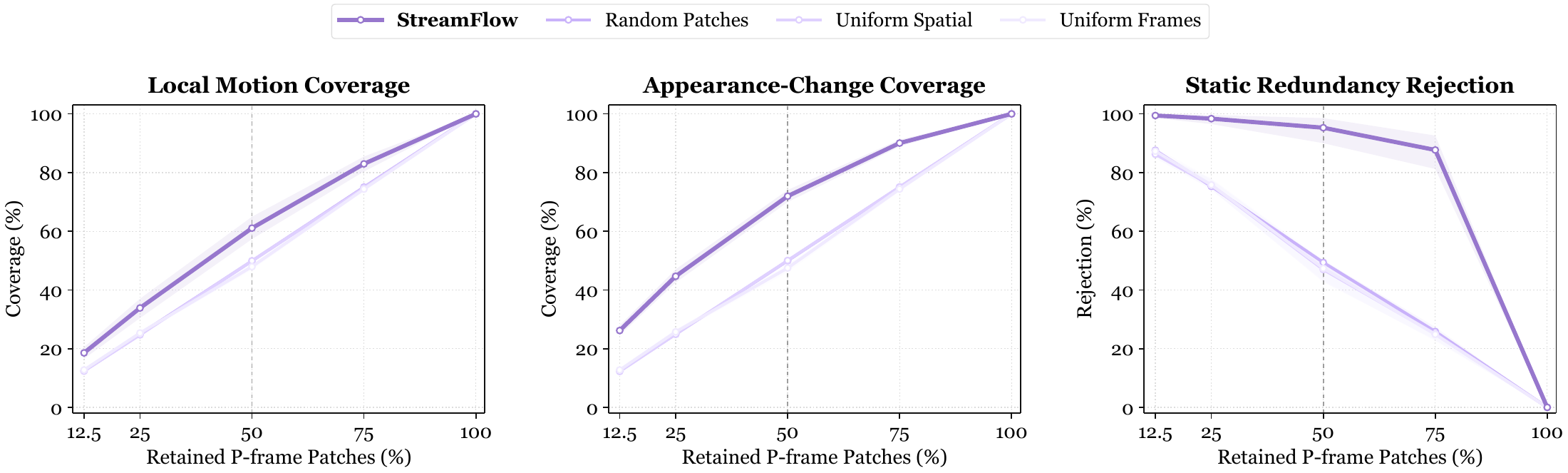}
    \captionof{figure}{Quantitative patch-retention analysis on RTVU. We vary the retained P-frame patch ratio while holding the budget identical across residual Top-K, random patches, uniform spatial selection, and uniform frame selection. Curves report video-level means, and shaded regions denote $95\%$ bootstrap confidence intervals. The dashed line marks the default retention ratio $\rho=0.50$.}
    \label{fig:supp_retention_ratio_statistics}
\end{minipage}
\par\medskip
\subsection{Efficiency Measurement Boundaries}\label{app:efficiency_boundaries}

We report efficiency under three explicitly separated measurement scopes on RTVU: the visual front end, the decoder, and the complete task. Vanilla and \ourmethod are evaluated on identical questions and processed records, with the same $16\times28$ pre-merge patch grid and greedy decoding; \ourmethod uses a P-frame retention ratio of $50\%$. Each entry in the main-paper efficiency table is the arithmetic mean of the corresponding per-example measurements. Given a Vanilla measurement $m_{\mathrm{V}}$ and a \ourmethod measurement $m_{\mathrm{S}}$, we compute the reported reduction as $(m_{\mathrm{V}}-m_{\mathrm{S}})/m_{\mathrm{V}}\times100\%$. \Cref{tab:supp_efficiency_boundaries} makes the accounting boundary of each scope explicit.

\par\smallskip
\noindent\begin{minipage}{\textwidth}
    \centering
    \small
    \setlength{\tabcolsep}{3.0pt}
    \renewcommand{\arraystretch}{1.08}
    \captionof{table}{Measurement boundaries for the efficiency results on RTVU. The three scopes provide complementary measurements of distinct inference stages.}
    \label{tab:supp_efficiency_boundaries}
    \begin{tabular}{@{}p{0.20\columnwidth}p{0.74\columnwidth}@{}}
        \toprule
        \textbf{Scope} & \textbf{Measurement boundary} \\
        \midrule
        Visual front end &
        \textbf{Includes:} residual scoring, grouped Top-$K$, sparse packing, host-to-device transfer, and ViT encoding.
        \textbf{Excludes:} disk reads, shared resizing, model loading, and compilation. \\
        Decoder &
        \textbf{Includes:} valid context length, backbone BF16 K/V cache, and isolated causal attention.
        \textbf{Excludes:} vision, projections, MLPs, compressor, retrieval, VAS, and other decoder operations. \\
        Complete task &
        \textbf{Includes:} input materialization, vision, memory compression, retrieval, VAS, post-insertion forward passes, and greedy decoding.
        \textbf{Excludes:} disk reads, model loading, compilation, frame arrival, and GOP waiting. \\
        \bottomrule
    \end{tabular}
\end{minipage}
\par\medskip

\paragraph{Metric Definitions.}
Pre-ViT patches denote the number of valid patches presented to the vision transformer before spatial merging, summed over the observed video prefix of each example. Context length includes all valid text tokens, raw visual tokens, and retrieved latent tokens supplied to the answering backbone. KV-cache memory accounts for the BF16 keys and values retained across all backbone layers. Visual-front-end latency and self-attention latency isolate their respective stages, whereas task latency measures the complete inference path defined below. Task peak memory is the maximum allocator-reported memory observed within the same complete-task boundary.

\paragraph{Visual-Front-End Measurement.}
Timing begins after the patch payloads and source images are resident in host memory. Vanilla packs and encodes every patch, whereas \ourmethod first performs raw-pixel residual scoring, grouped Top-$K$ selection, sparse packing, and host-to-device transfer before encoding the retained patches. All static shapes are precompiled before timing. We measure each input three times and report the mean of the per-input medians over RTVU. This protocol measures offline benchmark execution rather than per-frame service latency; consequently, the reported $50.0\%$ reduction already includes the selection and packing overhead introduced by \ourmethod.

\paragraph{Decoder Measurement.}
KV-cache memory includes only the retained keys and values, excluding model weights, temporary activations, and the compressor. For the isolated causal-attention benchmark, we pad sequences to $128$-token blocks, perform two warm-up runs followed by seven timed runs, and scale the median per-layer latency by the decoder depth. This scope excludes projections, MLPs, vision, retrieval, and VAS computation.

\paragraph{Complete-Task Measurement and VAS Overhead.}
Complete-task timing begins with the processed video records and question resident in host memory and ends when greedy decoding finishes. This boundary includes the visual front end, compressor, memory maintenance, retrieval, token-wise VAS computation, and every post-insertion forward pass. At each decoding step, VAS is accumulated over all attention heads in every decoder layer and reduced on device to per-layer scalars; full attention matrices are neither materialized nor transferred to the host. We synchronize device execution before stopping the timer. Unlike the isolated KV-cache metric, allocator-sampled peak memory covers the complete model state, temporary execution state, and compressor. Under this end-to-end boundary, \ourmethod reduces mean peak memory from $13.68$ to $10.80$ GiB ($21.1\%$) and mean task latency from $7.36$ to $3.65$ seconds ($50.4\%$).

Complete-task timing excludes disk video decoding, model loading, first-time compilation, and frame-arrival time, including the wall-clock delay required to collect a GOP at $1$ FPS.

\subsection{Case Studies}\label{app:long_video_cases}

\begin{figure*}[p]
    \centering
    \includegraphics[width=0.98\textwidth,height=0.82\textheight,keepaspectratio,trim=0 1bp 0 0,clip]{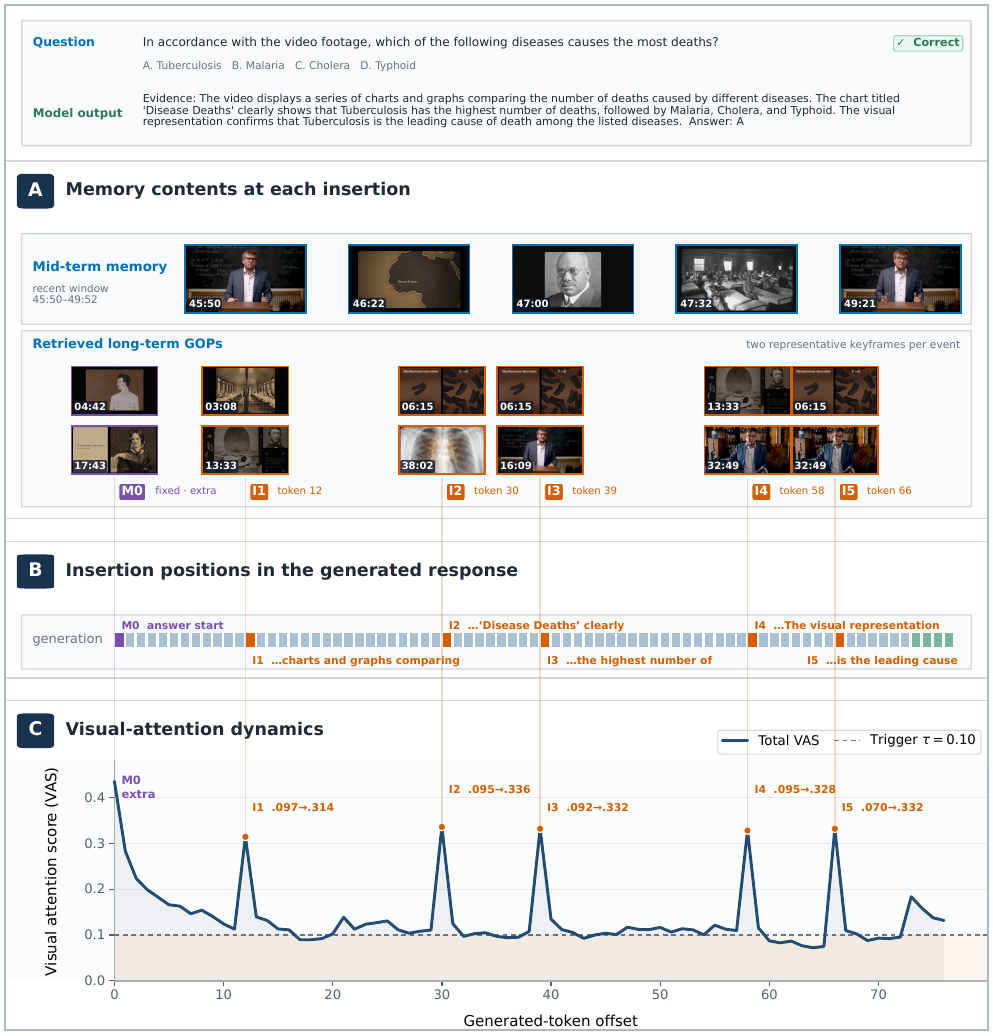}
    \caption{Token-aligned case study for an approximately $50$-minute VideoMME-Long sample. The top card shows the question and correct model output. \textbf{(A)} Mid-term memory and retrieved long-term GOPs. \textbf{(B)} Insertion positions during generation. \textbf{(C)} VAS dynamics across generated tokens.}
    \label{fig:supp_videomme_case_study}
\end{figure*}

\begin{figure*}[p]
    \centering
    \includegraphics[width=0.98\textwidth,height=0.82\textheight,keepaspectratio,trim=0 1bp 0 0,clip]{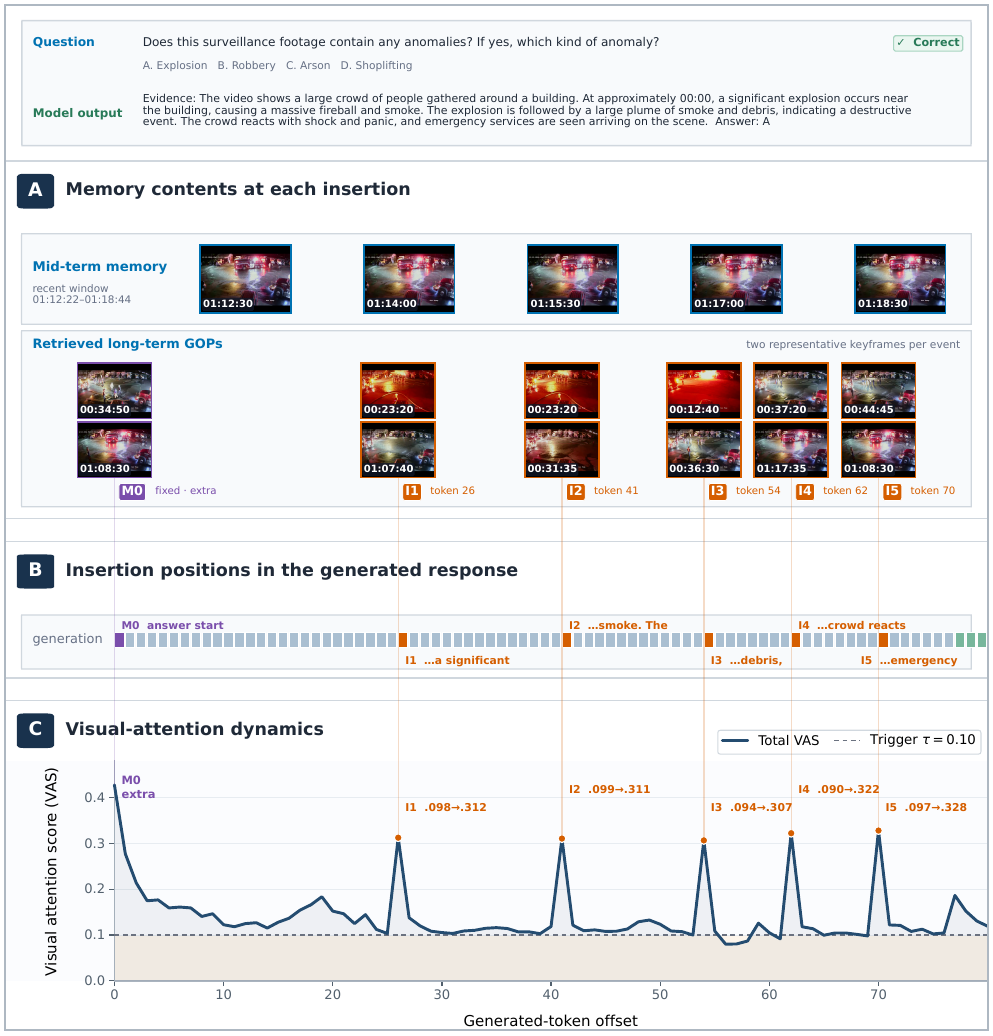}
    \caption{Token-aligned case study for an approximately $79$-minute MLVU anomaly-recognition sample. The top card shows the question and correct model output. \textbf{(A)} Mid-term memory and retrieved GOPs. \textbf{(B)} Insertion positions during generation. \textbf{(C)} VAS dynamics across generated tokens.}
    \label{fig:supp_mlvu_case_study}
\end{figure*}

\Cref{fig:supp_videomme_case_study,fig:supp_mlvu_case_study} align available memory and retrieved GOPs with answer generation and VAS trajectories in two representative cases. Together, they visualize how evolving answer prefixes guide retrieval and how dynamic insertion restores access to long-range visual evidence.

\paragraph{VideoMME-Long.}
\Cref{fig:supp_videomme_case_study} traces an approximately $50$-minute example. Mid-term memory covers the final four minutes, while long-term retrieval reaches earlier disease-history evidence, including the \emph{Mycobacterium tuberculosis} frame at $00{:}06{:}15$ and the lung X-ray at $00{:}38{:}02$.

As the response develops, five VAS-guided insertions raise visual attention from below $\tau=0.10$ to above $0.31$. The retrieved evidence spans distant parts of the video and progressively grounds the model's disease-focused reasoning in the relevant visual history.

\paragraph{MLVU.}
\Cref{fig:supp_mlvu_case_study} examines a $78.8$-minute surveillance video. Mid-term memory covers $01{:}12{:}22$--$01{:}18{:}44$, while retrieved GOPs revisit the fire, emergency vehicles, and responders from as early as $00{:}12{:}25$, supplying temporally distributed evidence for anomaly recognition.

Five dynamic insertions again lift VAS from below $0.10$ to above $0.30$ as the response identifies the explosion and accumulates supporting details about the fireball, smoke, debris, and arriving emergency services. StreamFlow consequently produces a complete evidence-grounded response and correctly selects ``A: Explosion.''

\end{document}